\pdfoutput=1

\documentclass[11pt]{article}

\usepackage[]{EMNLP2023}

\usepackage{times}
\usepackage{latexsym}

\usepackage[T1]{fontenc}

\usepackage[utf8]{inputenc}

\usepackage{microtype}

\usepackage{inconsolata}

\usepackage{graphicx}
\usepackage{float}
\usepackage{subcaption}
\usepackage{multirow}
\usepackage{makecell}
\usepackage{amsmath}
\usepackage{multicol}
\usepackage{booktabs}
\usepackage{enumitem}
\newcommand{\factor}[1]{{\fontfamily{cmtt}\selectfont #1}}

\usepackage{xcolor}         
\usepackage{colortbl}         
\usepackage{amssymb}  

\title{Mind the instructions: a holistic evaluation of consistency and interactions in prompt-based learning
}

\author{Lucas Weber \\
  University Pompeu Fabra \\
  \texttt{lucas.weber@upf.edu} \\\And
  Elia Bruni \\
  Osnabrück University \\
  \texttt{elia.bruni@gmail.com} \\\And
  Dieuwke Hupkes \\
  FAIR \\
  \texttt{dieuwkehupkes@meta.com}
}

\begin{document}
\maketitle
\begin{abstract}
Finding the best way of adapting pre-trained language models to a task is a big challenge in current NLP.
Just like the previous generation of \textit{task-tuned} models (TT), models that are adapted to tasks via in-context-learning (ICL) are robust in some setups but not in others.
Here, we present a detailed analysis of which design choices cause instabilities and inconsistencies in LLM predictions.
First, we show how spurious correlations between input distributions and labels -- a known issue in TT  models -- form only a minor problem for prompted models.
Then, we engage in a systematic, holistic evaluation of different factors that have been found to influence predictions in a prompting setup.
We test all possible combinations of a range of factors on both vanilla and instruction-tuned (IT) LLMs of different scale and statistically analyse the results to show which factors are the most influential, interactive or stable.
Our results show which factors can be used without precautions and which should be avoided or handled with care in most settings.

\end{abstract}

\section{Introduction}
Transfer learning from large-scale pre-trained language models is nowadays the standard approach to a wide range of NLP tasks. 
One of its great challenges is to optimally interface information that pre-trained language models accumulate in their parameters and adapt it to the task of interest \cite{zhou2023lima, ouyang2022training}. 
The standard approach for task adaptation has recently shifted from updating model parameters for a specific task (from here on task tuning or \emph{TT}) to using prompting-based methods based on in-context learning (from here on \emph{ICL}).
ICL can be subdivided into few-shot \citep[][]{brown2020language} or zero-shot inference \citep[primarily using instruction-tuned models ][]{wei2021finetuned}.
Both approaches offer certain benefits over TT: it eliminates costly, task-specific finetuning and provides greater flexibility, as a single model can be applied to many tasks.
However, ICL also currently yields overall weaker performance compared to task-tuning and is less stable and reliable on many benchmarks \citep[see, e.g.][]{bang2023multitask, ohmer2023evaluating, min2022rethinking,lu2021fantastically,zhao2021calibrate}. 

While for TT, much research has been conducted to understand weaknesses in the paradigm \citep[for an overview, see ][]{hupkes2023taxonomy}, the sources of instabilities in ICL remain nebulous.
Since ICL is more constrained (less data and no parameter updates), out-of-distribution generalisation has been suggested to be less of a problem \citep{awadalla2022exploring, si2023prompting}.
On the other hand, new frontiers emerge. 
For example, the format, order, or semantics of provided in-context examples can greatly influence learning outcomes, as does the proportion of labels in the context and the exact labels used \citep{liang2022holistic}.
Little is known, however, about how these factors interact \citep[work from ][suggests that they cannot be isolated]{wei2023larger, kim2022ground}; it is unclear which aspects are consistently beneficial, which vary across setups, and which are sensible to combine or decouple.
The volatility of the paradigm warrants more research into the reliability of different design choices.

In this paper, we conduct a detailed exploration of vanilla and instruction-tuned LLMs across various shifts and setups to understand their robustness.
We start with one of the prominent themes in robustness studies for TT models: robustness to spurious correlations between input and label distributions \citep{kavumba2019choosing, mccoy2019right, niven2019probing} and find that in ICL, spurious correlations do not have a significant impact on learning outcomes.

We go on to investigate ICL's sensitivity to other features of adaptation context, as well as the consistency of predictions across different design choices.
To do so, we conduct a large-scale grid search across various combinations of factors and statistically analyse the results to shed light on the inter-dependencies of different design choices.
We find that the exact in-context setup (the number of in-context examples, the distribution of in-context labels, or the type of instructions given in the context) has a surprisingly small but reliable impact on prediction outcomes.
On the other hand, the type of instructions used to query the target has, by far, the most significant impact on model behaviour. It is also the most volatile across settings, making it the most pivotal factor.

\section{Background and related work}
\label{sec:background}
In the following, we first briefly define TT and ICL and then cover known problems with model robustness.

\subsection{Task tuning and spurious correlations}
\label{subsec:background_task_tuning}
TT aligns a pre-trained model with a specific task by iteratively updating model parameters to minimise prediction loss on adaptation data.
In our definition here, TT does not include finetuning on more abstract objectives like instruction tuning \citep[IT; ][]{wei2021finetuned}.
TT models often fit spurious correlations between inputs and associated labels that are idiosyncratic artefacts to the specific dataset \citep{niven2019probing, kavumba2019choosing, mccoy2019right, geva2019we, poliak2018hypothesis, gururangan2018annotation, kavumba2022prompt} and do not align with the causal structure of the process that generated the data in `the real world' \citep{scholkopf2012causal}. 
Such adaptations \citep[sometimes also referred to as `shortcut solutions';][]{geirhos2020shortcut} usually fail as soon as the data distribution shifts between the adaptation and test phase. 
Pre-training improves robustness compared to task training from scratch \citep{hendrycks2019using, hendrycks2020pretrained}. 
However, the necessary posthoc task adaptation still overfits spurious correlations \citep{niven2019probing}.
An effective way to mitigate issues in task adaptation is to expose the model to counterexamples of spurious correlations \citep{kaushik2019learning}.

\subsection{In-context learning}
\label{subsec:background_ICL}
ICL describes the adaptation of a model to a task by inferring the task from the input given to the model. 
ICL can be subdivided into (1) few-shot learning, where in-context examples (consisting of input-output pairs) are given in the left-handed context of a tested input, and (2) zero-shot learning, referring to the case in which there are no examples.
In this paper, we investigate few-shot scenarios.

In contrast to TT, ICL is a considerably cheaper adaptation method as it does not require any parameter updates. 
\citet{akyurek2022learning} and \citet{garg2022can} show that adaptation of transformer models via ICL exhibits the same degree of expressivity as simple linear algorithms, small neural networks or decision trees. 
While ICL emerges spontaneously with increasing size of untuned LLMs \cite{brown2020language}, the ICL performance of such \emph{`vanilla'} LLMs lags behind the tuned state-of-the-art on almost all common NLP benchmarks \citep{liang2022holistic}.

Previous research has also shown that ICL is highly unstable. For example, the order of in-context examples \citep{lu2021fantastically}, the recency of certain labels in the context \citep{zhao2021calibrate} or the format of the prompt \citep{mishra2021reframing} as well as the distribution of training examples and the label space \citep{min2022rethinking} strongly influence model performance.
Curiously, whether the labels provided in the examples are *correct* is less important\citep{min2022rethinking}.
However, these findings are not uncontested: \citet{kim2022ground} paint a more differentiated picture, demonstrating that in-context input-label mapping \emph{does} matter, but that it depends on other factors such as model size or instruction verbosity.
Along a similar vein, \citet{wei2023larger} show that in-context learners can acquire new semantically non-sensical mappings from in-context examples if presented in a specific setup.

From this listing, we see that ICL entails many design choices, that task-unrelated design choices change prediction outcomes and that the effects of design choices do not exist in isolation.
The field is only beginning to understand the complex interplays of different prompting setups. 

\newcommand{\tabularwidth}{\textwidth}

\newcommand{\expone}{$\square$}  
\newcommand{\exptwo}{$\bigtriangleup$}

\begin{table*}[h!]
\centering
\renewcommand{\arraystretch}{1.1}         
\setlength{\tabcolsep}{0mm}         
\begin{tabular}{|p{\tabularwidth}<{\centering}|}         
\hline
               
\rowcolor{gray!60}               
\textbf{Motivation} \\               
\footnotesize
\begin{tabular}{p{0.25\tabularwidth}<{\centering} p{0.25\tabularwidth}<{\centering} p{0.25\tabularwidth}<{\centering} p{0.25\tabularwidth}<{\centering}}                        
\textit{Practical} & \textit{Cognitive} & \textit{Intrinsic} & \textit{Fairness}\\
\expone\hspace{0.8mm}	\exptwo\hspace{0.8mm}	
& 		
& \expone\hspace{0.8mm}	\exptwo\hspace{0.8mm}	
& 		

\vspace{2mm} \\
\end{tabular}\\
               
\rowcolor{gray!60}               
\textbf{Generalisation type} \\               
\footnotesize
\begin{tabular}{m{0.21\tabularwidth}<{\centering} m{0.2\tabularwidth}<{\centering} m{0.13\tabularwidth}<{\centering} m{0.13\tabularwidth}<{\centering} m{0.13\tabularwidth}<{\centering} m{0.19\tabularwidth}<{\centering} m{0.01\tabularwidth}<{\centering}}                   
\textit{Compositional} & \textit{Structural} & \textit{Cross Task} & \textit{Cross Language} & \textit{Cross Domain} & \textit{Robustness} & \\
& 		
& \expone\hspace{0.8mm}	
& 		
& 		
& \expone\hspace{0.8mm} \exptwo\hspace{0.8mm}
&
\end{tabular}\\
             
\rowcolor{gray!60}             
\textbf{Shift type} \\             
\footnotesize
\begin{tabular}{p{0.25\tabularwidth}<{\centering} p{0.25\tabularwidth}<{\centering} p{0.25\tabularwidth}<{\centering} p{0.24\tabularwidth}<{\centering} p{0.01\tabularwidth}<{\centering}}                        
\textit{Covariate} & \textit{Label} & \textit{Full} & \textit{Assumed} & \\  
\exptwo\hspace{0.8mm}	
& 		
& 		
& \expone\hspace{0.8mm}		
&
\vspace{2mm} \\
\end{tabular}\\
             
\rowcolor{gray!60}             
\textbf{Shift source} \\             
\footnotesize
\begin{tabular}{p{0.25\tabularwidth}<{\centering} p{0.25\tabularwidth}<{\centering} p{0.25\tabularwidth}<{\centering} p{0.25\tabularwidth}<{\centering}}                          
\textit{Naturally occuring} & \textit{Partitioned natural} & \textit{Generated shift} & \textit{Fully generated}\\
& \expone\hspace{0.8mm}		
& \exptwo\hspace{0.8mm}		
& 		

\vspace{2mm} \\
\end{tabular}\\
             
\rowcolor{gray!60}             
\textbf{Shift locus}\\             
\footnotesize
\begin{tabular}{p{0.25\tabularwidth}<{\centering} p{0.25\tabularwidth}<{\centering} p{0.25\tabularwidth}<{\centering} p{0.24\tabularwidth}<{\centering}p{0.01\tabularwidth}<{\centering}}                         
\textit{Train--test} & \textit{Finetune train--test} & \textit{Pretrain--train} & \textit{Pretrain--test} &\\
& \exptwo\hspace{0.8mm}	
& 		
& \expone\hspace{0.8mm}		
&
\end{tabular}\\

\hline
\end{tabular}
\caption{Our analyses, categorised according to the GenBench taxonomy \cite{hupkes2023taxonomy}. The token \exptwo\hspace{0.8mm} represents Experiment I and \expone\hspace{0.8mm} represents Experiment II.}
\label{tab:GenBenchEvalCard}
\end{table*}

\section{Experiment I: Robustness to spurious correlations}
\label{sec:spurious_correlations}
We clarify open questions about robustness of in-context learners by elucidating their sensitivity to factors to which they should be invariant (from here on \emph{invariance factors}).
First, we focus on one of the most prominent forms of non-robustness in TT models: susceptibility to spurious correlations between inputs and labels (see Section~\ref{subsec:background_task_tuning}).
In the first set of experiments, we test how different models behave when spurious correlations are contained in their adaptation data. 

\subsection{Setup}
\label{subsec:exp1_setup}
We here describe the datasets and models used to test sensitivity to spurious correlations.

\paragraph{Datasets}

\begin{table}[!ht]
\centering
\footnotesize
\begingroup
\setlength{\tabcolsep}{3pt}
\renewcommand{\arraystretch}{1}
\begin{tabular}{lll}
\hline
\textbf{Task} & \textbf{Base dataset} &  \textbf{Adversarial dataset}  \\
\hline

\rule{0pt}{2ex}NLI &MNLI \tiny{\citep{williams2017broad}} &  HANS \tiny{\citep{mccoy2019right}} \\ && ANLI \tiny{\citep{nie2019adversarial}} \\
PI &QQP \tiny{\citep{wang2017bilateral}} &  PAWS \tiny{\citep{zhang2019paws}}  \\ 
QA &SQuAD \tiny{\citep{rajpurkar2016squad}} & SQuAD adv. \tiny{\citep{jia2017adversarial}} \\ && adv. QA \tiny{\citep{bartolo2020beat}} \\ && SQuAD shifts \tiny{\citep{miller2020effect}} \\
\hline
\end{tabular}
\endgroup
\caption{Tasks and corresponding datasets.
}
\label{tab:datasets}
\end{table}
We use different common NLU datasets (from here on \emph{base datasets}) which are known to contain spurious correlations between input and label distributions \citep{gururangan2018annotation, geva2019we, poliak2018hypothesis}, as well as \emph{adversarial datasets} of the same tasks.
Adversarial datasets are designed to not contain the spurious correlations of the base datasets; then, they can be used to test whether models use short-cut solutions (for an overview see Table~\ref{tab:datasets}).
Our base datasets span three different types of NLU tasks: natural language inference (NLI), paraphrase identification (PI) and extractive question answering (QA). 
An overview can be found in Table~\ref{tab:datasets} and additional details about dataset properties and their construction in Appendix~\ref{app:dataset_details}.


\paragraph{Models}
Our first experiment compares TT models with models that perform tasks through ICL. 
For the latter, we consider two types of models: `\emph{vanilla}' LLMs, and LLMs that are tuned to follow instructions \citep[\emph{IT} see e.g.][]{wei2021finetuned, zhong2021adapting}.

For TT, we use models based on RoBERTa\textsubscript{BASE} and RoBERTa\textsubscript{LARGE} \citep{liu2019roberta}. 
If available, we reutilise finetuned versions of RoBERTa that have been open-sourced through the huggingface hub \citep{wolf2019huggingface}; if not available, we finetune the respective models ourselves (with training details in Appendix~\ref{app:ft_details}).

\begin{table}[h!]
\centering
\footnotesize
\begin{tabular}{ll}
\hline
\textbf{Type of learning} & \textbf{Model}\\
\hline
\rule{0pt}{2ex}TT & RoBERTa-base \\
 & RoBERTa-large \\
ICL + vanilla 
 & LLaMA 7B, 13B, 30B, 65B \\
ICL + Instruction-tuning & Alpaca 7B, 13B, 30B, 65B\\
\hline
\end{tabular}
\caption{Adaptation types and the respective models, as used in Section~\ref{sec:spurious_correlations}. We use the same ICL models in Section~\ref{sec:holistic_eval}.
}

\label{tab:learning-models}
\end{table}

\begin{figure*}[h!]
  \begin{subfigure}[h]{0.70\linewidth}
  \centering
    \includegraphics[width=\linewidth]{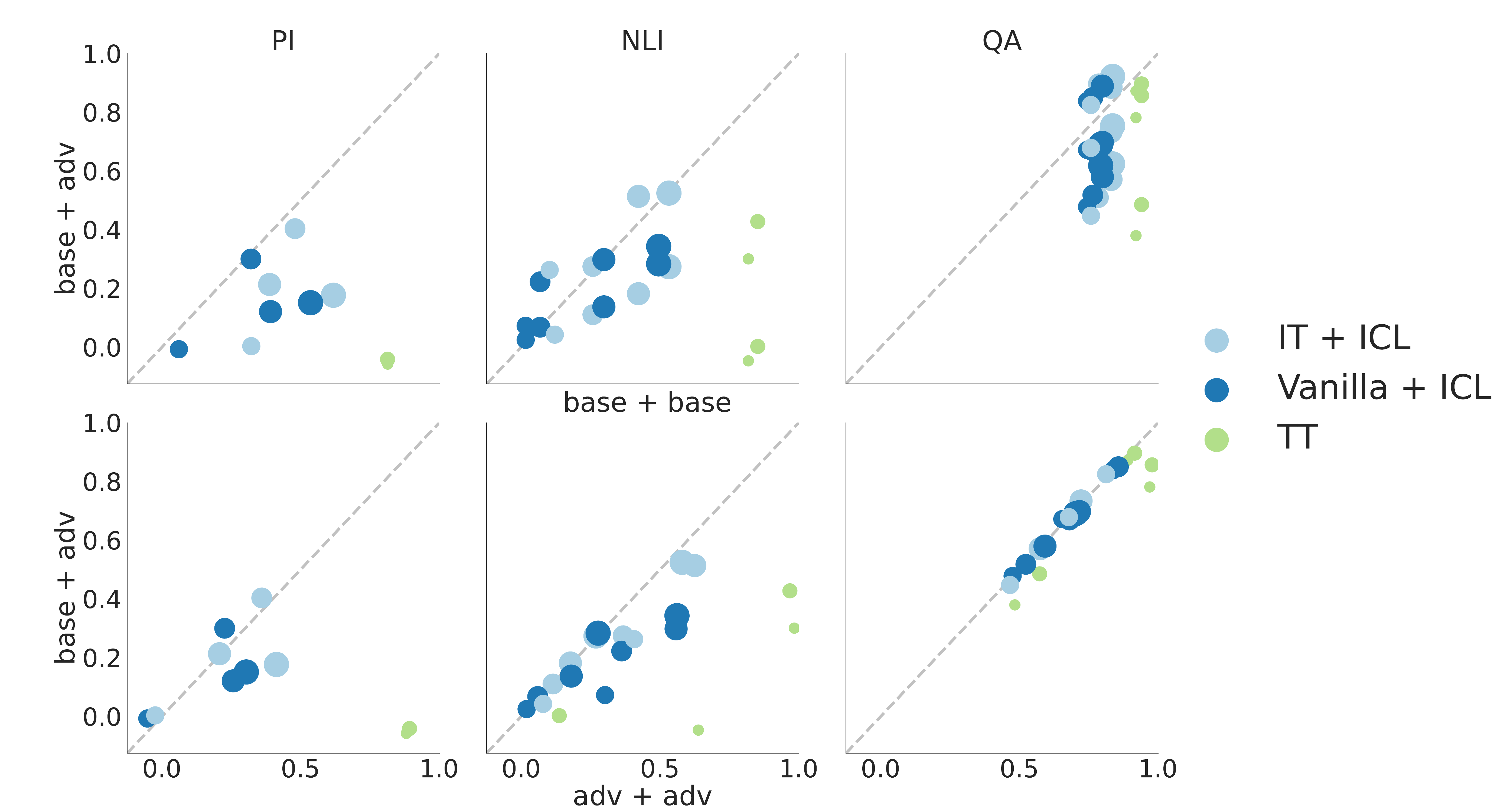}
    \caption{}
    \label{subfig:scatter_spurious}
  \end{subfigure}
  \begin{subfigure}[h]{0.21\linewidth}
  \vspace{0.2cm}
  \centering
    \includegraphics[width=\linewidth]{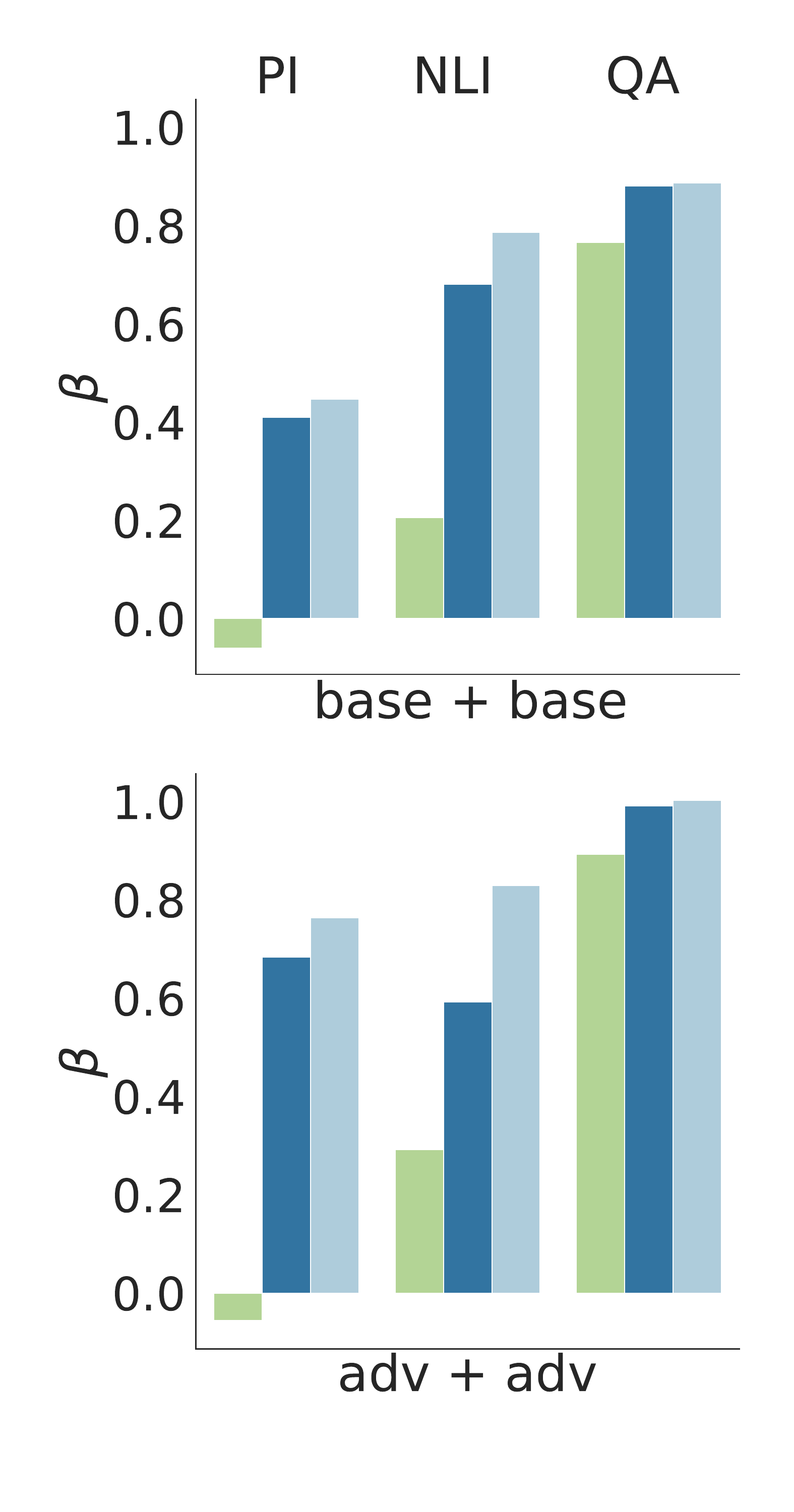}
    \caption{}
    \label{subfig:log_reg_b_values}
  \end{subfigure}
  \caption{
Figure (a) shows the f1-scores of different models -- normalised for random accuracy -- on different datasets when adapted via base or adversarial data. 
On each y-axis, we plot accuracy under distributional shift (\emph{base + adv}) while on each x-axis there is no shift (\emph{base + base} or \emph{adv + adv}). 
Each column shows a different type of task.
Marker size represents model size and colour represents the type of task adaptation.
Dots close to the diagonal indicate invariance to the adaptation data and therefore robust generalisation, while dots in the bottom right indicate sensitivity to spurious correlations.
Figure (b) shows the $\beta$-parameter of the linear regression (fixed intercept) on the data of Figure (a).
We fit a linear regression for each task and adaptation type separately.
Values close to 0 indicate very strong sensitivity to adaptation data, while values close to 1 indicate no sensitivity.
}
  \label{fig:scatters}
\end{figure*}

Our vanilla LLMs consist of the series of LLaMA models \cite[7B, 13B, 33B, 65B;][]{touvron2023llama}.
The IT counterparts are the freely available Alpaca models, which are based on the same LLaMA models but are additionally fine-tuned via low-rank adaptation \citep[LoRA;][]{hu2021lora} on the Alpaca self-instruct dataset \citep{alpaca, wang2022self}.
We run all models using mixed-precision decomposition as described by \citet{dettmers2022llm}. 
For an overview of all used models, see Table~\ref{tab:learning-models}.

\paragraph{Evaluation}
We evaluate ICL models by concatenating the target example $x$ with $k$ labelled in-context examples and greedily decoding from the probability distribution over possible labels $y\in\mathcal{C}$ using $argmax_{y\in\mathcal{C}}P(y|x_1, y_1...x_k, y_k, x)$ where $\mathcal{C}$ is the set of possible labels.
Every data point $x$ is wrapped by an \emph{instruction} template that explains the task the model should solve in natural language. 
The label space $\mathcal{C}$ is determined by the type of instruction template and can differ across templates.
We mitigate the influences of potential confounds like the template format, the order of $(x_i, y_i)$, imbalanced distribution of $y_i$ or the semantics of $x_i$ by a pseudo-random sampling $x_i$ for every new model inference.
Our sampling of $x_i$ ensures that the in-context labels $y_i$ are balanced over all possible labels \citep[similar to][inter alia]{wei2023larger, brown2020language}.
Moreover, we use multiple instruction templates sourced from FLAN \citep{wei2021finetuned} to avoid systematic bias. 

\subsection{Results}
\label{subsec:exp1_results}
We first evaluate the capacity of different models to robustly generalise from adaptation data to test data.
In the taxonomy of generalisation capabilities, this constitutes a \textit{covariate shift} between the adaptation data (finetuning data in TT and in-context data in ICL) and the test data \citep[compare GenBench; ][]{hupkes2023taxonomy}.
The corresponding GenBench evaluation card can be found in Table~\ref{tab:GenBenchEvalCard}.

\paragraph{Base data in-context}
First, we adapt the TT and ICL models on the base data and then compare their performance between the base data and the respective adversarial counterparts. 
If an approach is robust to spurious correlations in the adaptation data (which are the fine-tuning data or in-context examples, respectively), it should perform approximately equally on the base dataset and the adversarial dataset. 
We relate both scores in the first row of Figure~\ref{fig:scatters}. 

Results from in-context learners land generally closer to the diagonal, hence indicating -- despite overall weaker performance -- that they are more robust to the spurious correlations in their adaptation data.
To quantify this visual result, we fit a linear regression model on the data presented in the scatterplot in Figure~\ref{subfig:scatter_spurious} (hence, predict the adversarial- from the base accuracies) with the intercept fixed at $\beta_0 = 0$. 
The coefficient $\beta_1$ can then be interpreted as a degree of robustness to the different adaptation data, with $\beta_1 = 1$ indicating complete robustness and $\beta_1 = 0$ complete reliance on non-generalisable patterns in the base data. 
The $\beta_1$ values for different adaptation types can be found in the top row of Figure~\ref{subfig:log_reg_b_values}.
The $\beta_1$ values across all tasks are significantly closer to the parity value of 1 for ICL models than for TT models, with IT models having the edge over vanilla models.

Our results demonstrate that ICL models are much less sensitive to spurious correlations in their adaptation data than TT models.
However, the fact that ICL models do not reach the parity value of 1 means that gains on adversarial data are smaller compared to gains on the base data.
This suggests that ICL may still be mildly sensitive to spurious correlations, or, alternatively, that the adversarial datasets used are simply inherently more difficult, resulting in lower performances compared to the base data\footnote{An illustrative example of the base data being easier: adversarial QA contains only a single answer alternative while squad contains three.}. 
We will further explore this question in the next experiment.

\paragraph{Adversarial data in-context}
As a follow-up experiment, we consider what happens when the adaptation data contains adversarial examples. 
As those examples do not contain the same spurious correlations, models cannot overfit them \citep{kaushik2019learning}. 
This should not make a difference for models that are robust to spurious correlations, but we expect a performance drop between these two conditions for models that learned solutions that exploited those correlations.
As we are now evaluating the adversarial data points in both scenarios, we eliminate the potential impact of the dataset difficulty on the scores.
In the second row of Figure~\ref{fig:scatters}, we plot performances with base adaptation examples in the context against the performance with adversarial adaptation data, noting that ICL models are mostly unaffected by adaptation data type while TT models land far underneath the diagonal again.
A regression analysis shows almost all $\beta$-values of ICL models moving closer to parity, showing us how the dataset difficulty impacted the results. 
However, even without the effect of dataset difficulty on the $\beta$-values, they are still not quite equal to 1, suggesting that the type of adaptation data \textit{has} a small influence on ICL learners.

\begin{figure*}[h!]
  \centering
    \begin{subfigure}[h]{0.345\linewidth}
    \includegraphics[width=\linewidth]{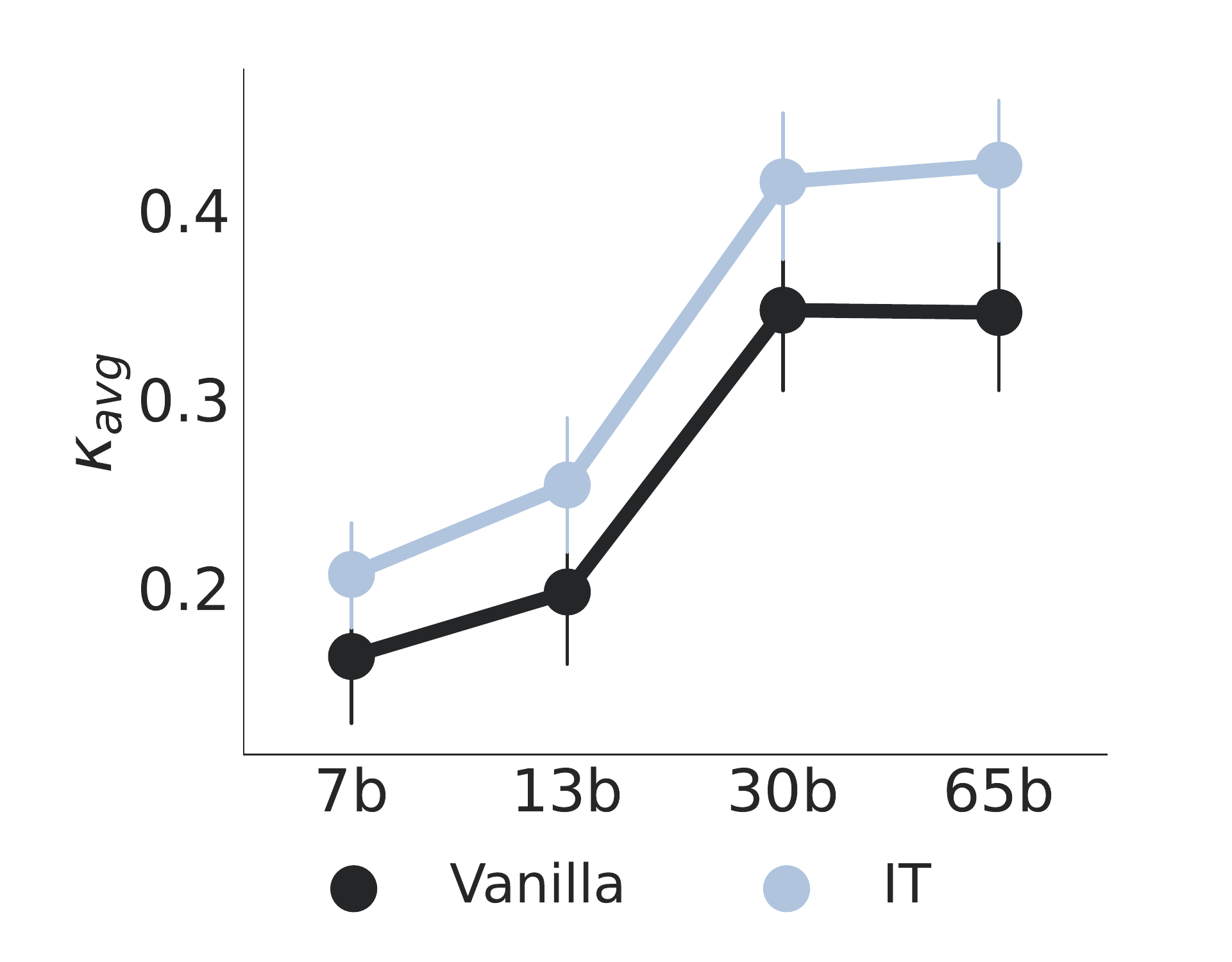}
    \caption{}
    \label{subfig:consistency_templates_per_model}
  \end{subfigure}
  \begin{subfigure}[h]{0.315\linewidth}
    \includegraphics[width=\linewidth]{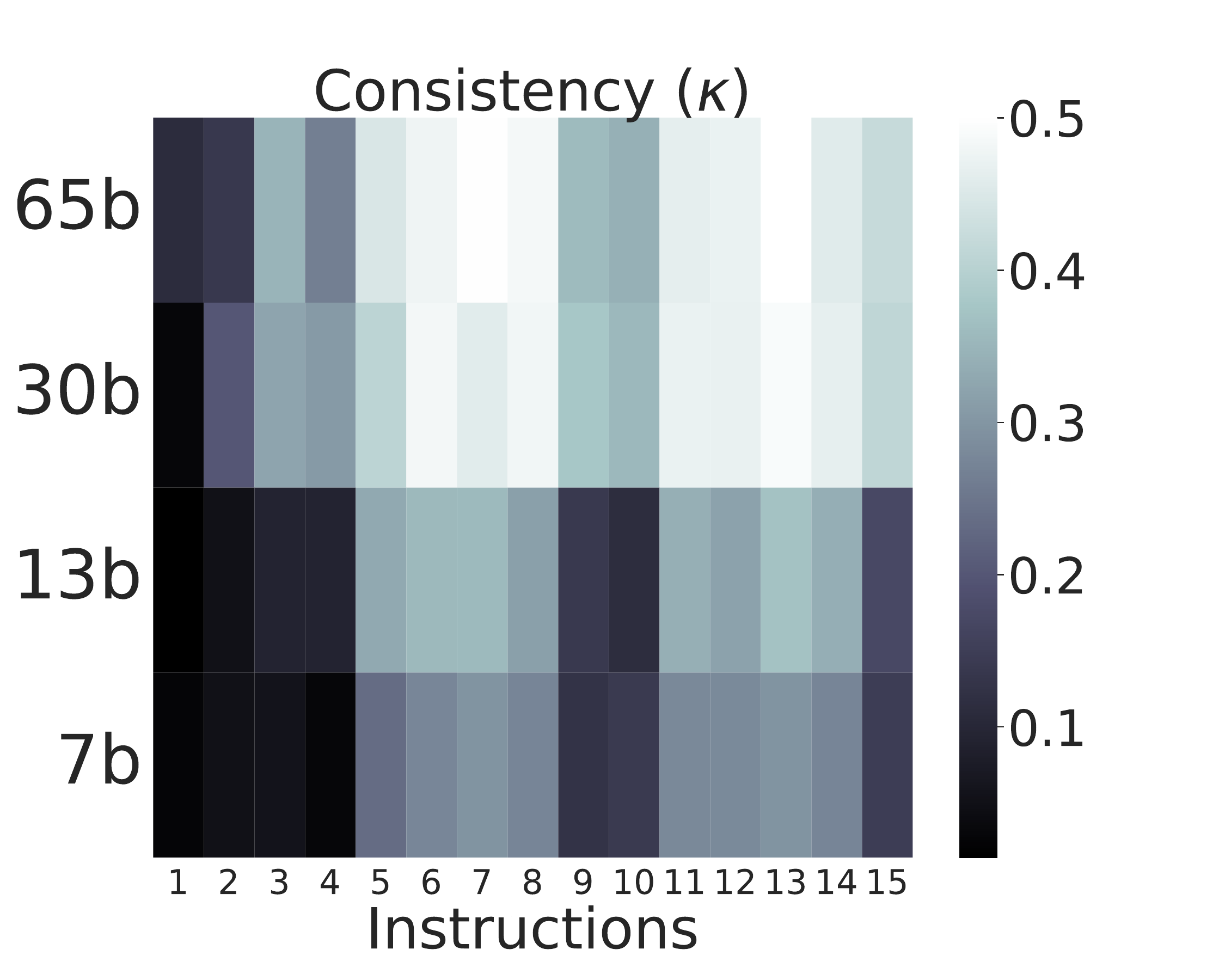}
    \caption{}
    \label{subfig:consistency_templates_per_template}
  \end{subfigure}
  \begin{subfigure}[h]{0.315\linewidth}
    \includegraphics[width=\linewidth]{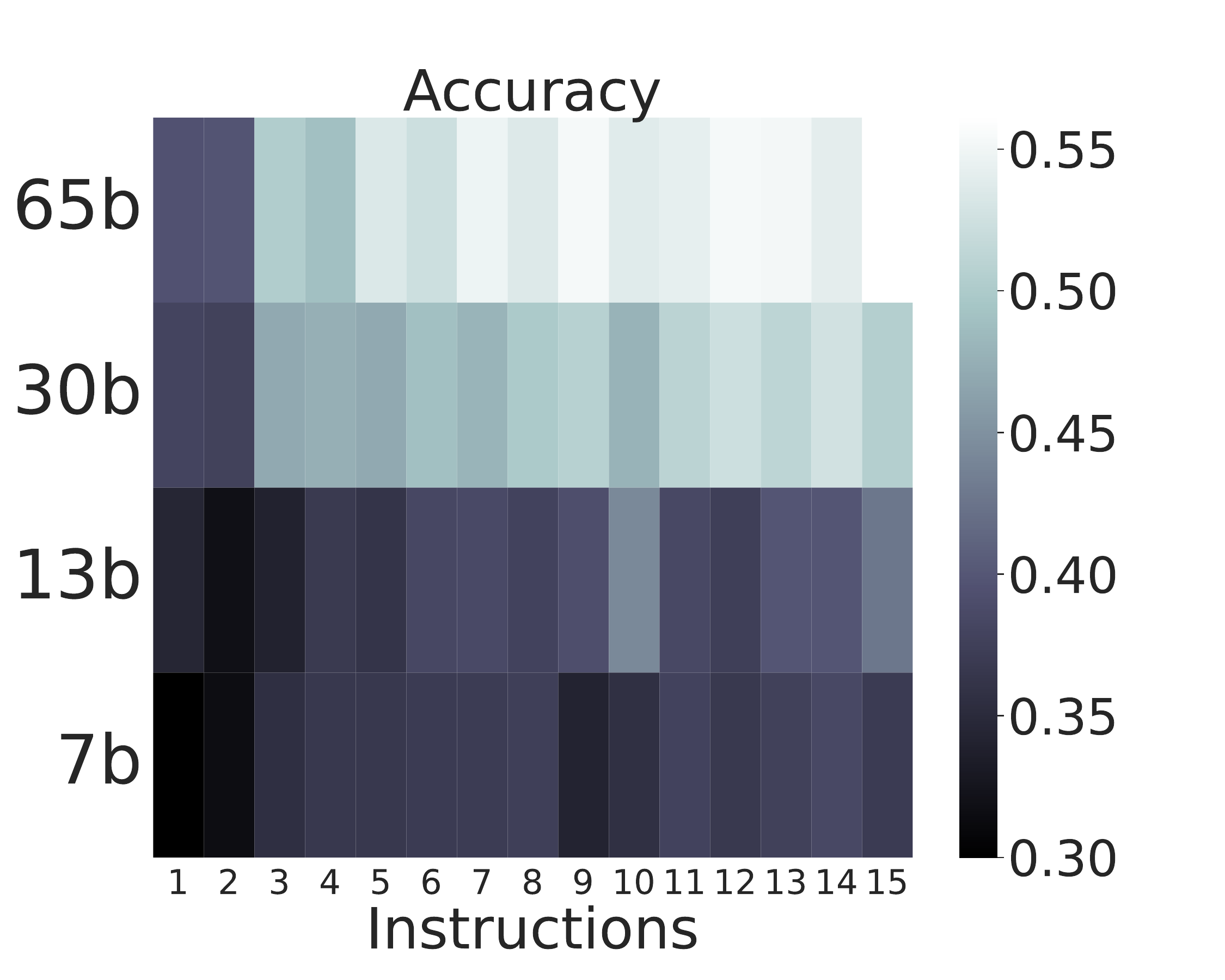}
    \caption{}
    \label{subfig:heatmap_performance_templates}
  \end{subfigure}
  \caption{Figure (a) shows the consistency of a model when used with all 15 different P3 instructions, in an otherwise fixed setup. A value of 1 indicates perfect agreement (all templates produce the same prediction); 
  Figure (b) shows how consistent individual instructions are with all other instructions. A value of 0 indicates a complete change of predictions while a value of 1 indicates perfect agreement; Figure (c) shows the respective accuracies of the instructions in Figure (b).}
  \label{fig:diversity_predictions}
\end{figure*}

\section{Experiment II: Consistency evaluation in ICL}
\label{sec:holistic_eval}
In the previous section, we saw that the robustness of in-context learners is likely influenced more by other factors than by spurious correlations in the in-context data. 
Although previous studies have reported the susceptibilities of LLMs to various factors, the impact of different design decisions and their interactions in the context of ICL robustness has not been systematically evaluated.
Here, we test the effects of an extensive range of these factors on prediction outcomes in consistency and accuracy.


\subsection{Experimental details}
\label{subsec:exp2_setup}
For all of the following experiments, we use promptsource templates \cite[P3;][]{bach2022promptsource} and the ANLI dataset \cite{nie2019adversarial}.
We continue to use the models and the evaluation procedure from Section~\ref{sec:spurious_correlations} (excluding the TT models). 
The following briefly describes the factors we consider in our analysis.

\subsubsection{Factors}
We distinguish two types of factors. Firstly, we consider factors that constitute interventions to improve consistency and performance, which we call \textbf{variance factors}\footnote{For detailed explanations on the different factors, we refer to Appendix~\ref{app:factors_details}.} or $\lambda_{var}$ for short. We expect a model to \emph{change} their response when we change the value of those factors: 
    \begin{description}[noitemsep]
        \item[Size]
        We consider models with 7B, 13B, 30B and 65B learnable parameters.
        \item[Instruction tuning] Whether models are instruction-tuned or not (`vanilla' models).
        \item[Calibration] Whether model outputs are calibrated using `content-free prompts' following \citet{zhao2021calibrate}.
        \item[n-shots] Whether there are many ($k$ = 5) or few ($k$ = 2) in-context examples in the prompt.
        \item[Instruction quality] Whether instructions belong to one of two groups of semantically equivalent but \emph{differently performing} instruction templates (high- vs. low-performing; more details in Section~\ref{subsubsec:probing_templates}).
        \item[Balanced labels] Whether examples with labels are balanced across all possible classes in the context or use randomly sampled examples.
        \end{description}
Secondly, we consider factors from which we want a model to \emph{not change} their response (or `be robust to') when we change their value. We will call these \textbf{invariance factors} or $\lambda_{inv}$: 
    \begin{description}[noitemsep]
    \item[Cross-templates] Whether in-context instructions are drawn randomly from all available instruction templates or use the same instructions as for the target.
    \item[Cross-task] Whether another classification task (QQP) is used as in-context examples or the same task as the target task (ANLI) is used.
    \item[Instructions] Different semantically equivalent target instructions that \emph{perform similarly} (more details in Section~\ref{subsubsec:probing_templates}).
    \item[One label] Whether in-context examples have only a \textit{single} randomly selected label or diverse labels.
    \end{description}

\begin{figure*}[h!]
  \centering
    \includegraphics[width=\linewidth]{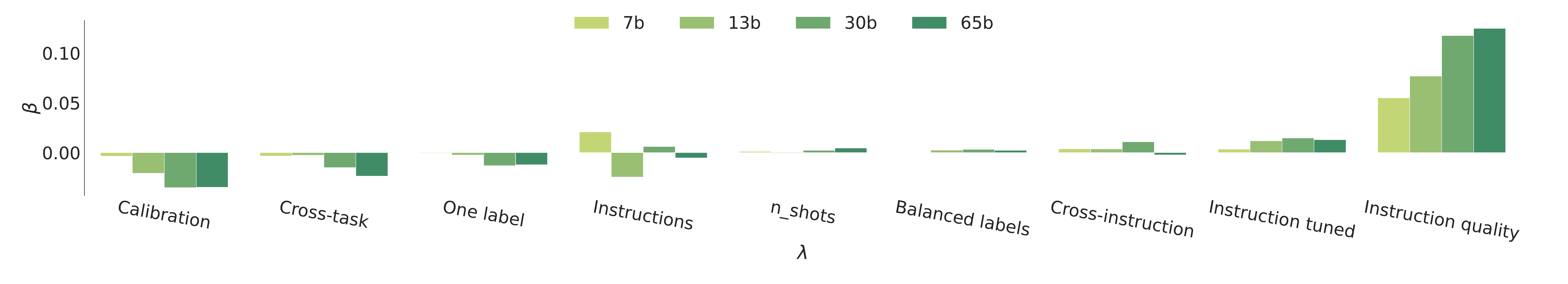}
  \caption{The $\beta$-values of the main effects of each individual factor across many different runs. The values can be directly interpreted as `\emph{expected accuracy gain/loss}' when a factor is present compared to when it is absent.}
  \label{fig:main_effects}
\end{figure*}
    
Combining the above factors results in 1536 setups.
We evaluate each of these constellations using the same subset of 600 data points\footnote{We found 600 examples to yield sufficiently similar results to evaluating the whole dataset, tested on a small subset of setups.} that we draw uniformly from either of the ANLI validation sets.
In-context examples are drawn at random from the respective training sets.

\subsubsection{Analysis methods} 
\label{subsec:exp2_analysis_method}
Our analysis entails two steps:

1. Main effects: how much does a single factor impact consistency and the accuracy across many setups?

2. Interactions: when we disentangle the main effects, do we find systematic interactions across pairs or triplets of factors?

\paragraph{Main effects} To evaluate the main effect of each factor $\lambda$, we employ linear regression to predict the accuracy of a model based on $\lambda$, considering all possible combinations of the remaining factors. 
The regression model is formulated as $Acc = \beta_1\lambda + \beta_0$. 
The coefficient $\beta_1$ represents the main effect of a specific $\lambda$, approximating the average change in accuracy across all possible setups given $\lambda$. 
We also fit the intercept $\beta_0$, but won't interpret it.

\paragraph{Interactions} We analyse interactions by fitting a factorial ANOVA considering the effect of all possible 2- and 3-way interactions\footnote{We exclude the \emph{instructions} factor because the independence of \emph{instruction quality} is not given.
Moreover, we adapt the significance levels via Bonferroni correction for multiple comparisons ($\alpha < 0,00059$) and show only significant interactions.} of factors on the accuracy of predictions. 
We then count the number of significant interactions every factor maintains with other factors.
A larger number of interactions suggests that a factor is volatile, i.e. it changes the predictions depending on the overall setup.
Further, as the factors have been chosen to be orthogonal and should not influence each other.
On the other hand, if factors are not interacting, we can interpret their main effects directly.

\subsubsection{Consistency metrics}
\label{subsubsec:consistency_metric}

We measure the consistency of model predictions using Cohen's $\kappa$ \citep[][]{cohen1960coefficient}, a measure of interrater agreement adjusted for agreement by chance.
The metric $\kappa$ equals 1 if two (or more) sets of predictions perfectly align while agreement by chance results in $\kappa$ equalling 0. 
In our case, we calculate $\kappa$ to compare the predictions of a model before and after we change the value of a factor $\lambda$ (e.g. if all labels in-context are the same or if they are not; see \factor{One label}) across all possible setups.
We make the metric less dependent on the accuracy of a model by calculating $\kappa$ only on the subset of predictions that have been correctly predicted in either of the two cases.

\subsubsection{Probing instructions} 
\label{subsubsec:probing_templates}
To find a set of high- and low-performing instructions for the \factor{instruction quality} factor, we run a preliminary analysis where we probe model behaviour in response to all 15 available P3 ANLI instructions.
We assess the performance of different instructions based on accuracy and consistency.

We first get a general picture of each model's average consistency $\kappa_{avg}$ across all templates.
We find that $\kappa_{avg}$ increases with the number of parameters and is overall higher when a model has been instruction tuned (Figure~\ref{subfig:consistency_templates_per_model}).

We then consider the consistency of each individual instruction and find a congruent pattern of consistency across all models (Figure~\ref{subfig:consistency_templates_per_template}) that corresponds generally to the accuracy scores of the same instructions (compare Figure~\ref{subfig:heatmap_performance_templates}).
Interestingly, we also find two groups of high-accuracy instructions making very different predictions (see the consistency scores of 9, 10 and 15 vs. rest).
Based on these observations, we choose the two highest- and lowest-performing instructions to constitute the \factor{instruction quality} factor and templates 14 and 15 as realisations of the \factor{instructions} factor.

\subsection{Results}
\label{subsec:exp2_results}

We evaluate the models on all possible combinations of $\lambda_{var}$ and $\lambda_{inv}$.
Appendix~\ref{app:distribution_results} shows the distribution of accuracy scores across all runs for different models.
The wide spread of scores is striking: large models score from below chance to up to 67\% accuracy, depending on the overall setup.
This extreme variability underlines the importance of better understanding the impact of different design decisions and prediction consistency in ICL. 
The subsequent section comprehensively summarises the results of our statistical analysis.

\subsubsection{Main effects} 
\label{subsubsec:exp2_res_main_effects}
The main effects separated by model size are shown in Figure~\ref{fig:main_effects}, illustrating each factor's impact in isolation.

\paragraph{Variance factors}
The variance factors we chose are generally thought to improve accuracy and, hence, should have positive main effects.
We find two out of five \emph{variance factors} significantly improve performance on average, from which \factor{instruction quality} stands out as the most influential factor across all model sizes. 
Similarly, we find that \factor{instruction tuning} is consistently beneficial while \factor{balancing} the in-context labels and the number of in-context examples (\factor{n-shots}) have on average positive but small and non-significant effects.
Surprisingly, \factor{calibration} harms rather than helps performance for all but our smallest model.

\paragraph{Invariance factors}
Different from variance factors, invariance factors are chosen such that they should not influence a robust model's predictions.
Accordingly, the main effects should be optimally close to 0.
We find that models are generally robust to having varied instructions in-context (\factor{cross-instruction}), or even having a slightly positive effect.
This is intriguing, as this factor entails considerable changes to the in-context setup, and we previously saw how the type of \emph{target instructions} (in \factor{instruction quality}) plays a major role. 
Further, we identify vulnerabilities of large models to the factors \factor{cross-task} and \factor{one label}.
The ambivalent effect of the \factor{instructions} factor suggests high volatility across similarly performing instructions (i.e. different instructions perform differently for different models and setups).

These main effects give us a general idea of the tendencies of factors.
To better understand all main effects, we will investigate interactions in Section~\ref{subsubsec:exp2_res_main_effects}.

\begin{figure}[h!]
  \centering
    \includegraphics[width=0.99\linewidth]{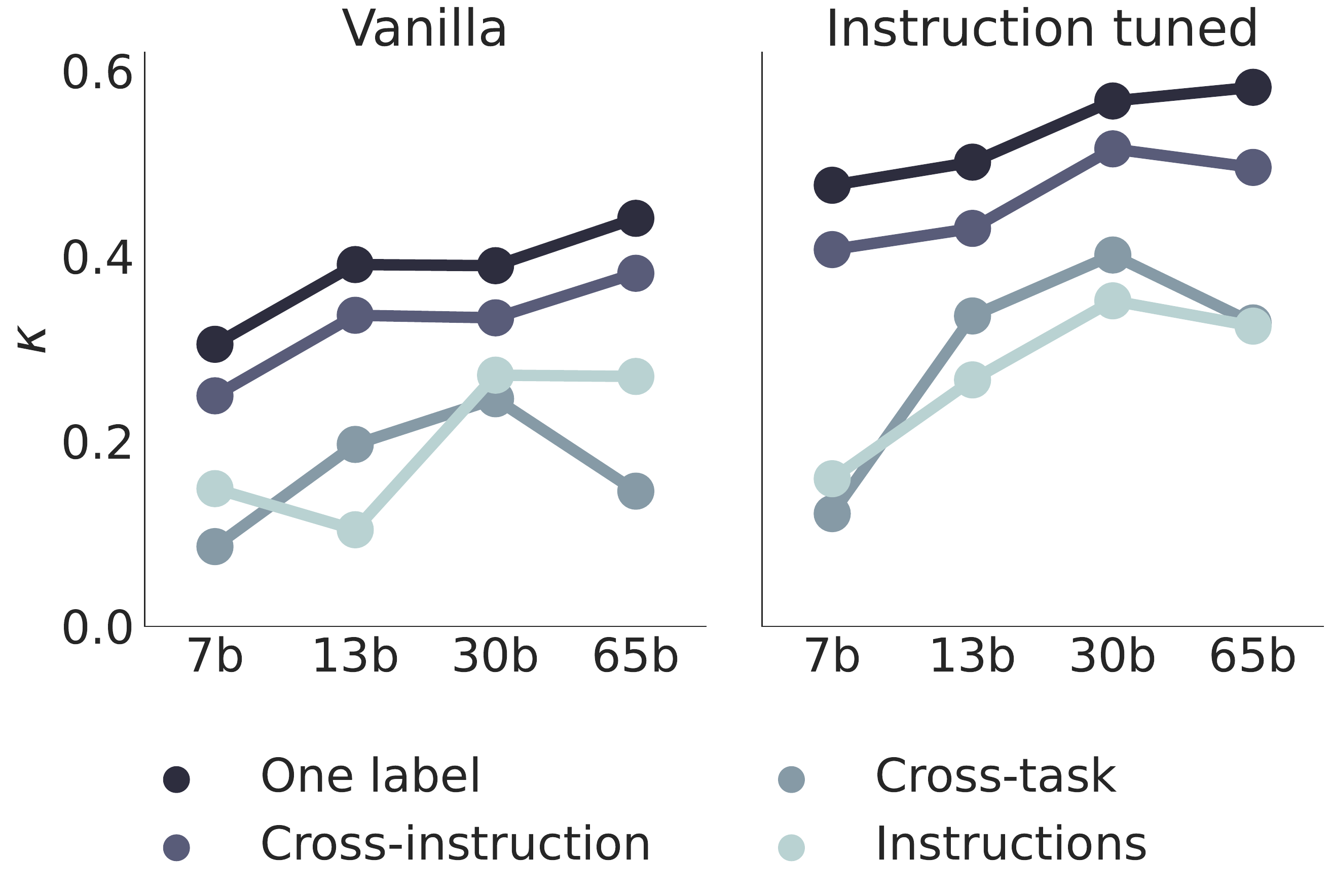}
  \caption{The consistency values when a specific factor is present or not across all other setups. A value of 0 indicates a complete change of predictions while a value of 1 indicates perfect agreement (i.e. a low value indicates that a model is not robust to a change in a specific factor).}
  \label{fig:consistency_lambdas}
\end{figure}

\paragraph{Consistency of invariance factors}
Additionally to a factor's impact on accuracy, we also compute the prediction consistency $\kappa$ of the factors (as defined in Section~\ref{subsubsec:consistency_metric}).
To do so, we calculate the agreement of predictions when a factor is present with when it is absent.
This way, the value of $\kappa$ shows us the degree of robustness of a model to an invariance factor by quantifying the degree of prediction change caused by that factor.
Figure~\ref{fig:consistency_lambdas} shows how robustness increases with size and instruction tuning.
The very low $\kappa$ scores for the detrimental \factor{cross-task} factor come as no surprise, while low scores in the \factor{instructions} factor corroborate the previous suspicion that instructions are highly volatile: if we change the type of used \factor{instructions}, the predictions across a lot of setups change.

\begin{figure*}[h!]
\centering
  \begin{subfigure}[h]{0.40\linewidth}
  \centering
    \includegraphics[width=\linewidth]{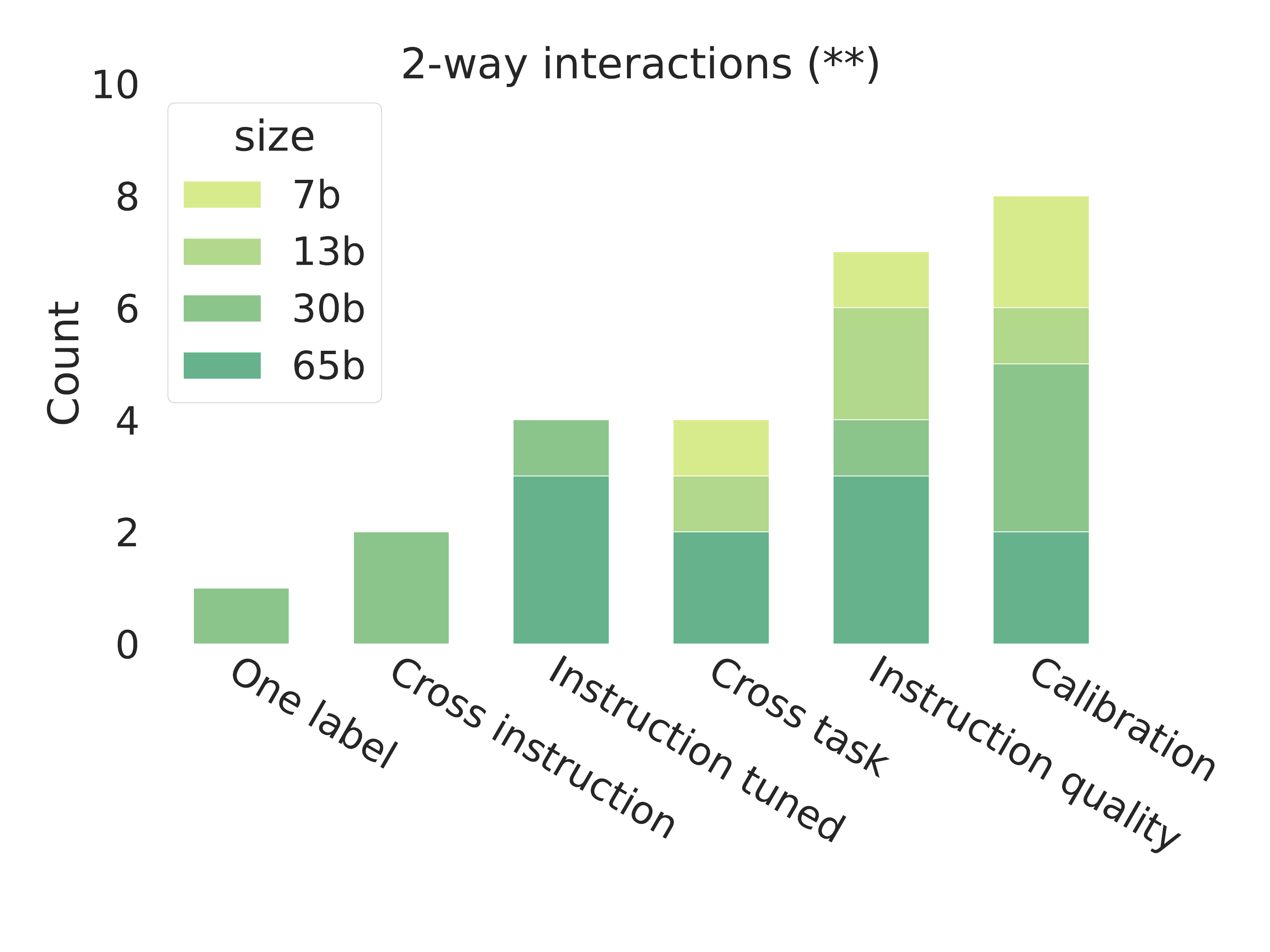}
    \label{subfig:hist_2way_interaction}
  \end{subfigure}
  \begin{subfigure}[h]{0.40\linewidth}
  \centering
    \includegraphics[width=\linewidth]{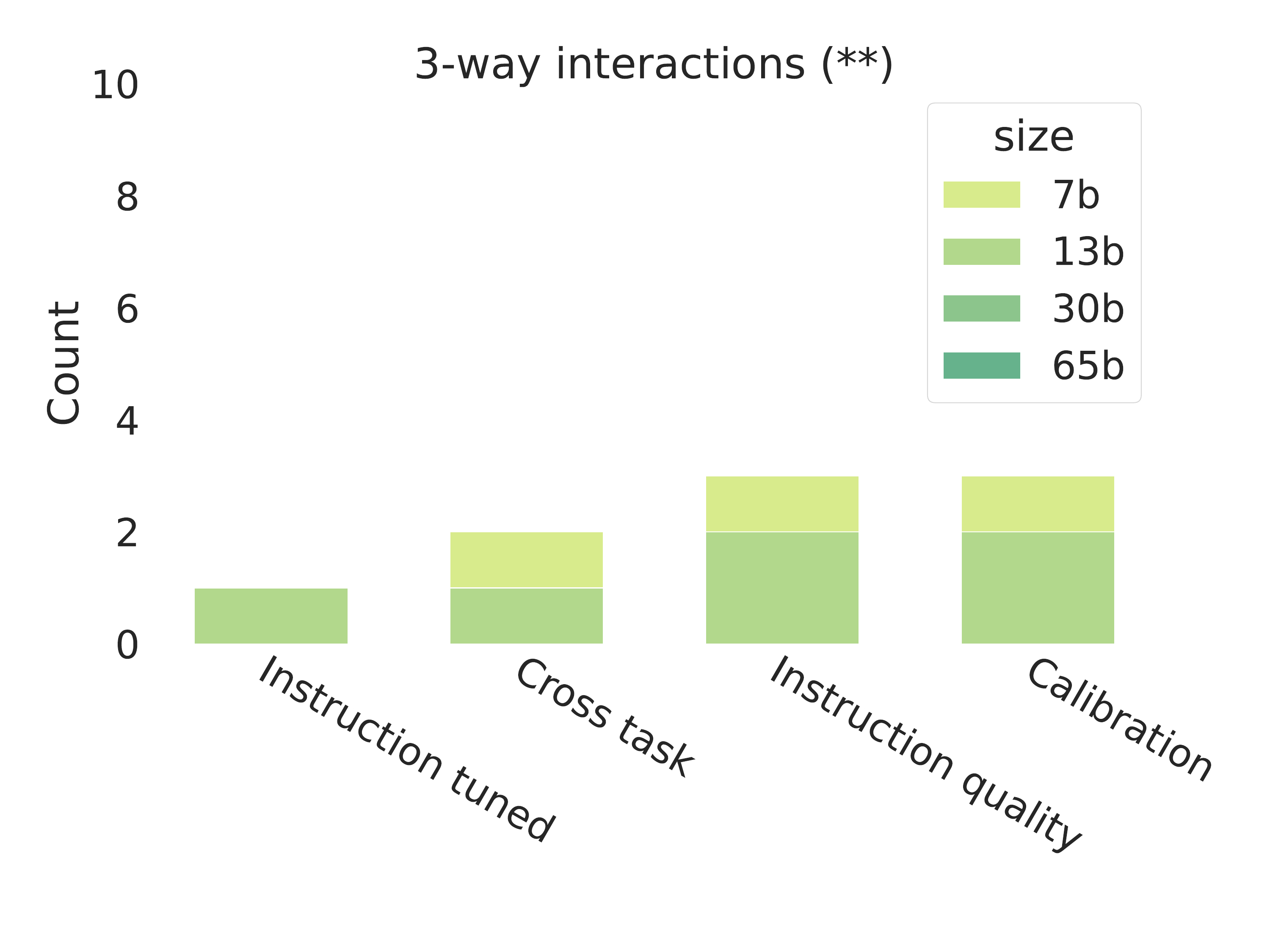}
    \label{subfig:hist_3way_interaction}
  \end{subfigure}
  \caption{The number of interactions per factor with other factors. A large number of interactions means that the outcome of a change in these factors depends on a lot of other variables.}
  \label{fig:interactions}
\end{figure*}

\subsubsection{Interactions} 
\label{subsubsec:exp2_res_interactions}
The main effects give us a good idea of the general direction of the impact of a single factor.
However, the main effects do not tell the whole story: 
consider the case in which factor \factor{A} improves performance if it is paired with factor \factor{B}, but performance deteriorates when paired with \factor{C}. 
\factor{A}'s overall main effect might be close to zero even though it influences certain settings.
To better understand the impact of each factor, we will have to investigate its interactions.

We determine interactions following the procedure described in Section~\ref{subsec:exp2_analysis_method}.
Figure~\ref{fig:interactions} shows the number of interactions that each factor maintains. 
A general observation is that large models tend to have simpler 2-way interactions, while smaller models tend to have more complex 3-way interactions.

\paragraph{Highly interactive factors}
The most important factor of \factor{instruction quality} maintains many interactions. 
Hence, many other factors change predictions depending on the used instruction template.
We find a similar effect for the \factor{instructions} factor\footnote{We fit another ANOVA excluding \factor{instruction quality} while keeping \factor{instructions} as a factor to ensure that the effect is not only due to large performance differences between the two realisations of instruction quality. We find similarly strong interactions for the \factor{instructions} factor (see Appendix~\ref{app:interactions_instructions}).}.
This demonstrates the intricacy of the formulation of instructions: 
the \factor{instruction quality} has the largest positive impact on prediction outcomes, but at the same time, the instructions are highly interactive and volatile, with their the effects of many other factors depending on it.

Otherwise, we observe that \factor{calibration} is the most volatile, with eight significant interactions with other factors.
The previously observed main effect has to be seen in this perspective: \factor{calibration} is not generally detrimental, but its effects depend very much on the setup in which it is used. 
For example, we find on closer inspection that \factor{calibration} leads to the highest overall accuracies for the 7B parameter models when presented with specific \factor{instructions} and paraphrase identification in-context examples (\factor{cross-task}). 

\paragraph{Low interactive factors}
On the other end of the spectrum, we find that factors like the number of in-context examples (\factor{n-shots}), the \factor{balancing} of in-context labels or using just \factor{one label} have little to no interactions at all.
Conveniently, there are no ambiguities for these factors and we can therefore interpret their main effects directly, as they are most likely to be stable across setups.
For example, suppose it is possible to increase the number of examples in the context. 
In that case, we can reliably expect small gains in accuracy without the danger of otherwise interfering with the learning process.
Similarly, balancing labels leads to reliable small improvements and having just a single label in the context reliably reduces accuracy for large models.


\section{Discussion}
\label{discussion}
We will first summarise the findings of this paper and then discuss their implications.

\paragraph{Findings}
We saw in Section~\ref{sec:spurious_correlations} how spurious correlations do not influence predictions in ICL in a relevant manner as they did previously in TT. 
This, however, does not resolve the problem of robustness: depending on the setup, ICL accuracy in our experiments differs up to 40\%, as other factors in the setup become pivotally important.
We here conducted a comprehensive analysis of the influence of different setups on the consistency of predictions in ICL models.
Considering different setups, well-chosen \factor{instructions} promise the largest performance gains across many setups. 
At the same time, they are among the most volatile factors of all and highly sensitive to the setting in which they are used.
On the other hand, factors that relate to the exact organisation of the in-context examples, such as the label distribution or in-context instructions (\factor{cross-instructions}), have surprisingly small impacts.
Other factors like \factor{n-shots} -- among others -- are not interactive, which makes them much easier to handle: their expected gain or loss should, in most cases, correspond to our observed main effects.
Across all of our experiments, we also find the general tendency that larger numbers of model parameters and instruction tuning are beneficial for model consistency across many settings.

\paragraph{Implications and future research}
What do these findings imply?
As we have seen, inconsistency is a severe concern in ICL, and we here contribute to narrowing down its sources.
Unlike previously in TT, concentrating on spurious correlations is not vital for ICL robustness and investigating design choices concerned with in-context examples (i.e. the exact few-shot setting) promises to be less impactful or mostly dependent on other setup factors.
Instead, our findings suggest that the exact phrasing of instruction templates is pivotally important.
To get hold of inconsistent predictions in ICL, finding the exact properties of instructions that so strongly influence model predictions is a sensible next step (potentially with a similar methodology as it is presented here).
Insights into the impact of instruction properties can help us to find the source of inconsistencies and avoid them in production, while they can also contribute to the theoretical understanding of in-context learning which is currently still under investigation.
While our analysis focused on the few-shot setting, it also significantly impacts the increasingly popular zero-shot learning, as instructions are central in that setting.

For model deployment, our findings demand caution as minor changes to certain parts of prompts (e.g. the instructions) can change the performance of the general setup. 
This is especially true for employing smaller, untuned models.
A consistent finding across all our experiments is that instruction tuning improves consistency and robustness to irrelevant factors across all setups.
Therefore, we advocate for the use of tuned models to improve robustness.

\section{Conclusion}
\label{conclusion}
We here analysed robustness and variability in the recent learning paradigm of ICL, showing that they are generally different from in task-tuning.
By using a methodology that covers a wide range of potential prompt design decisions, we show which factors actually matter in prompt design and how these factors influence each other.

\section*{Limitations and Acknowledgements}
For a discussion of the limitations of our work and the acknowledgements, we refer to Appendix~\ref{app:limitations} and Appendix~\ref{app:acknowledgements}, respectively.

\bibliography{anthology,custom}

\begin{thebibliography}{49}
\expandafter\ifx\csname natexlab\endcsname\relax\def\natexlab#1{#1}\fi

\bibitem[{Aky{\"u}rek et~al.(2022)Aky{\"u}rek, Schuurmans, Andreas, Ma, and
  Zhou}]{akyurek2022learning}
Ekin Aky{\"u}rek, Dale Schuurmans, Jacob Andreas, Tengyu Ma, and Denny Zhou.
  2022.
\newblock \href {https://arxiv.org/abs/2211.15661} {What learning algorithm is
  in-context learning? investigations with linear models}.
\newblock \emph{ArXiv preprint}, abs/2211.15661.

\bibitem[{Awadalla et~al.(2022)Awadalla, Wortsman, Ilharco, Min, Magnusson,
  Hajishirzi, and Schmidt}]{awadalla2022exploring}
Anas Awadalla, Mitchell Wortsman, Gabriel Ilharco, Sewon Min, Ian Magnusson,
  Hannaneh Hajishirzi, and Ludwig Schmidt. 2022.
\newblock \href {https://aclanthology.org/2022.findings-emnlp.441} {Exploring
  the landscape of distributional robustness for question answering models}.
\newblock In \emph{Findings of the Association for Computational Linguistics:
  EMNLP 2022}, pages 5971--5987, Abu Dhabi, United Arab Emirates. Association
  for Computational Linguistics.

\bibitem[{Bach et~al.(2022)Bach, Sanh, Yong, Webson, Raffel, Nayak, Sharma,
  Kim, Bari, Fevry, Alyafeai, Dey, Santilli, Sun, Ben-david, Xu, Chhablani,
  Wang, Fries, Al-shaibani, Sharma, Thakker, Almubarak, Tang, Radev, Jiang, and
  Rush}]{bach2022promptsource}
Stephen Bach, Victor Sanh, Zheng~Xin Yong, Albert Webson, Colin Raffel,
  Nihal~V. Nayak, Abheesht Sharma, Taewoon Kim, M~Saiful Bari, Thibault Fevry,
  Zaid Alyafeai, Manan Dey, Andrea Santilli, Zhiqing Sun, Srulik Ben-david,
  Canwen Xu, Gunjan Chhablani, Han Wang, Jason Fries, Maged Al-shaibani, Shanya
  Sharma, Urmish Thakker, Khalid Almubarak, Xiangru Tang, Dragomir Radev, Mike
  Tian-jian Jiang, and Alexander Rush. 2022.
\newblock \href {https://doi.org/10.18653/v1/2022.acl-demo.9}
  {{P}rompt{S}ource: An integrated development environment and repository for
  natural language prompts}.
\newblock In \emph{Proceedings of the 60th Annual Meeting of the Association
  for Computational Linguistics: System Demonstrations}, pages 93--104, Dublin,
  Ireland. Association for Computational Linguistics.

\bibitem[{Bang et~al.(2023)Bang, Cahyawijaya, Lee, Dai, Su, Wilie, Lovenia, Ji,
  Yu, Chung et~al.}]{bang2023multitask}
Yejin Bang, Samuel Cahyawijaya, Nayeon Lee, Wenliang Dai, Dan Su, Bryan Wilie,
  Holy Lovenia, Ziwei Ji, Tiezheng Yu, Willy Chung, et~al. 2023.
\newblock \href {https://arxiv.org/abs/2302.04023} {A multitask, multilingual,
  multimodal evaluation of chatgpt on reasoning, hallucination, and
  interactivity}.
\newblock \emph{ArXiv preprint}, abs/2302.04023.

\bibitem[{Bartolo et~al.(2020)Bartolo, Roberts, Welbl, Riedel, and
  Stenetorp}]{bartolo2020beat}
Max Bartolo, Alastair Roberts, Johannes Welbl, Sebastian Riedel, and Pontus
  Stenetorp. 2020.
\newblock \href {https://doi.org/10.1162/tacl_a_00338} {Beat the {AI}:
  Investigating adversarial human annotation for reading comprehension}.
\newblock \emph{Transactions of the Association for Computational Linguistics},
  8:662--678.

\bibitem[{Brown et~al.(2020)Brown, Mann, Ryder, Subbiah, Kaplan, Dhariwal,
  Neelakantan, Shyam, Sastry, Askell, Agarwal, Herbert{-}Voss, Krueger,
  Henighan, Child, Ramesh, Ziegler, Wu, Winter, Hesse, Chen, Sigler, Litwin,
  Gray, Chess, Clark, Berner, McCandlish, Radford, Sutskever, and
  Amodei}]{brown2020language}
Tom~B. Brown, Benjamin Mann, Nick Ryder, Melanie Subbiah, Jared Kaplan,
  Prafulla Dhariwal, Arvind Neelakantan, Pranav Shyam, Girish Sastry, Amanda
  Askell, Sandhini Agarwal, Ariel Herbert{-}Voss, Gretchen Krueger, Tom
  Henighan, Rewon Child, Aditya Ramesh, Daniel~M. Ziegler, Jeffrey Wu, Clemens
  Winter, Christopher Hesse, Mark Chen, Eric Sigler, Mateusz Litwin, Scott
  Gray, Benjamin Chess, Jack Clark, Christopher Berner, Sam McCandlish, Alec
  Radford, Ilya Sutskever, and Dario Amodei. 2020.
\newblock \href
  {https://proceedings.neurips.cc/paper/2020/hash/1457c0d6bfcb4967418bfb8ac142f64a-Abstract.html}
  {Language models are few-shot learners}.
\newblock In \emph{Advances in Neural Information Processing Systems 33: Annual
  Conference on Neural Information Processing Systems 2020, NeurIPS 2020,
  December 6-12, 2020, virtual}.

\bibitem[{Cohen(1960)}]{cohen1960coefficient}
Jacob Cohen. 1960.
\newblock A coefficient of agreement for nominal scales.
\newblock \emph{Educational and psychological measurement}, 20(1):37--46.

\bibitem[{Dettmers et~al.(2022)Dettmers, Lewis, Belkada, and
  Zettlemoyer}]{dettmers2022llm}
Tim Dettmers, Mike Lewis, Younes Belkada, and Luke Zettlemoyer. 2022.
\newblock \href {https://arxiv.org/abs/2208.07339} {Llm. int8 (): 8-bit matrix
  multiplication for transformers at scale}.
\newblock \emph{ArXiv preprint}, abs/2208.07339.

\bibitem[{Garg et~al.(2022)Garg, Tsipras, Liang, and Valiant}]{garg2022can}
Shivam Garg, Dimitris Tsipras, Percy~S Liang, and Gregory Valiant. 2022.
\newblock What can transformers learn in-context? a case study of simple
  function classes.
\newblock \emph{Advances in Neural Information Processing Systems},
  35:30583--30598.

\bibitem[{Geirhos et~al.(2020)Geirhos, Jacobsen, Michaelis, Zemel, Brendel,
  Bethge, and Wichmann}]{geirhos2020shortcut}
Robert Geirhos, J{\"o}rn-Henrik Jacobsen, Claudio Michaelis, Richard Zemel,
  Wieland Brendel, Matthias Bethge, and Felix~A Wichmann. 2020.
\newblock Shortcut learning in deep neural networks.
\newblock \emph{Nature Machine Intelligence}, 2(11):665--673.

\bibitem[{Geva et~al.(2019)Geva, Goldberg, and Berant}]{geva2019we}
Mor Geva, Yoav Goldberg, and Jonathan Berant. 2019.
\newblock \href {https://doi.org/10.18653/v1/D19-1107} {Are we modeling the
  task or the annotator? an investigation of annotator bias in natural language
  understanding datasets}.
\newblock In \emph{Proceedings of the 2019 Conference on Empirical Methods in
  Natural Language Processing and the 9th International Joint Conference on
  Natural Language Processing (EMNLP-IJCNLP)}, pages 1161--1166, Hong Kong,
  China. Association for Computational Linguistics.

\bibitem[{Gururangan et~al.(2018)Gururangan, Swayamdipta, Levy, Schwartz,
  Bowman, and Smith}]{gururangan2018annotation}
Suchin Gururangan, Swabha Swayamdipta, Omer Levy, Roy Schwartz, Samuel Bowman,
  and Noah~A. Smith. 2018.
\newblock \href {https://doi.org/10.18653/v1/N18-2017} {Annotation artifacts in
  natural language inference data}.
\newblock In \emph{Proceedings of the 2018 Conference of the North {A}merican
  Chapter of the Association for Computational Linguistics: Human Language
  Technologies, Volume 2 (Short Papers)}, pages 107--112, New Orleans,
  Louisiana. Association for Computational Linguistics.

\bibitem[{Hendrycks et~al.(2019)Hendrycks, Lee, and
  Mazeika}]{hendrycks2019using}
Dan Hendrycks, Kimin Lee, and Mantas Mazeika. 2019.
\newblock \href {http://proceedings.mlr.press/v97/hendrycks19a.html} {Using
  pre-training can improve model robustness and uncertainty}.
\newblock In \emph{Proceedings of the 36th International Conference on Machine
  Learning, {ICML} 2019, 9-15 June 2019, Long Beach, California, {USA}},
  volume~97 of \emph{Proceedings of Machine Learning Research}, pages
  2712--2721. {PMLR}.

\bibitem[{Hendrycks et~al.(2020)Hendrycks, Liu, Wallace, Dziedzic, Krishnan,
  and Song}]{hendrycks2020pretrained}
Dan Hendrycks, Xiaoyuan Liu, Eric Wallace, Adam Dziedzic, Rishabh Krishnan, and
  Dawn Song. 2020.
\newblock \href {https://doi.org/10.18653/v1/2020.acl-main.244} {Pretrained
  transformers improve out-of-distribution robustness}.
\newblock In \emph{Proceedings of the 58th Annual Meeting of the Association
  for Computational Linguistics}, pages 2744--2751, Online. Association for
  Computational Linguistics.

\bibitem[{Hu et~al.(2022)Hu, Shen, Wallis, Allen{-}Zhu, Li, Wang, Wang, and
  Chen}]{hu2021lora}
Edward~J. Hu, Yelong Shen, Phillip Wallis, Zeyuan Allen{-}Zhu, Yuanzhi Li,
  Shean Wang, Lu~Wang, and Weizhu Chen. 2022.
\newblock \href {https://openreview.net/forum?id=nZeVKeeFYf9} {Lora: Low-rank
  adaptation of large language models}.
\newblock In \emph{The Tenth International Conference on Learning
  Representations, {ICLR} 2022, Virtual Event, April 25-29, 2022}.
  OpenReview.net.

\bibitem[{Hupkes et~al.(2023)Hupkes, Giulianelli, Dankers
  et~al.}]{hupkes2023taxonomy}
D.~Hupkes, M.~Giulianelli, V.~Dankers, et~al. 2023.
\newblock \href {https://doi.org/10.1038/s42256-023-00729-y} {A taxonomy and
  review of generalization research in nlp}.
\newblock \emph{Nature Machine Intelligence}, 5:1161--1174.

\bibitem[{Jia and Liang(2017)}]{jia2017adversarial}
Robin Jia and Percy Liang. 2017.
\newblock \href {https://doi.org/10.18653/v1/D17-1215} {Adversarial examples
  for evaluating reading comprehension systems}.
\newblock In \emph{Proceedings of the 2017 Conference on Empirical Methods in
  Natural Language Processing}, pages 2021--2031, Copenhagen, Denmark.
  Association for Computational Linguistics.

\bibitem[{Kaushik et~al.(2020)Kaushik, Hovy, and Lipton}]{kaushik2019learning}
Divyansh Kaushik, Eduard~H. Hovy, and Zachary~Chase Lipton. 2020.
\newblock \href {https://openreview.net/forum?id=Sklgs0NFvr} {Learning the
  difference that makes {A} difference with counterfactually-augmented data}.
\newblock In \emph{8th International Conference on Learning Representations,
  {ICLR} 2020, Addis Ababa, Ethiopia, April 26-30, 2020}. OpenReview.net.

\bibitem[{Kavumba et~al.(2019)Kavumba, Inoue, Heinzerling, Singh, Reisert, and
  Inui}]{kavumba2019choosing}
Pride Kavumba, Naoya Inoue, Benjamin Heinzerling, Keshav Singh, Paul Reisert,
  and Kentaro Inui. 2019.
\newblock \href {https://doi.org/10.18653/v1/D19-6004} {When choosing plausible
  alternatives, clever hans can be clever}.
\newblock In \emph{Proceedings of the First Workshop on Commonsense Inference
  in Natural Language Processing}, pages 33--42, Hong Kong, China. Association
  for Computational Linguistics.

\bibitem[{Kavumba et~al.(2022)Kavumba, Takahashi, and Oda}]{kavumba2022prompt}
Pride Kavumba, Ryo Takahashi, and Yusuke Oda. 2022.
\newblock \href {https://doi.org/10.18653/v1/2022.acl-long.166} {Are
  prompt-based models clueless?}
\newblock In \emph{Proceedings of the 60th Annual Meeting of the Association
  for Computational Linguistics (Volume 1: Long Papers)}, pages 2333--2352,
  Dublin, Ireland. Association for Computational Linguistics.

\bibitem[{Liang et~al.(2022)Liang, Bommasani, Lee, Tsipras, Soylu, Yasunaga,
  Zhang, Narayanan, Wu, Kumar et~al.}]{liang2022holistic}
Percy Liang, Rishi Bommasani, Tony Lee, Dimitris Tsipras, Dilara Soylu,
  Michihiro Yasunaga, Yian Zhang, Deepak Narayanan, Yuhuai Wu, Ananya Kumar,
  et~al. 2022.
\newblock \href {https://arxiv.org/abs/2211.09110} {Holistic evaluation of
  language models}.
\newblock \emph{ArXiv preprint}, abs/2211.09110.

\bibitem[{Liu et~al.(2019)Liu, Ott, Goyal, Du, Joshi, Chen, Levy, Lewis,
  Zettlemoyer, and Stoyanov}]{liu2019roberta}
Yinhan Liu, Myle Ott, Naman Goyal, Jingfei Du, Mandar Joshi, Danqi Chen, Omer
  Levy, Mike Lewis, Luke Zettlemoyer, and Veselin Stoyanov. 2019.
\newblock \href {https://arxiv.org/abs/1907.11692} {Roberta: A robustly
  optimized bert pretraining approach}.
\newblock \emph{ArXiv preprint}, abs/1907.11692.

\bibitem[{Lu et~al.(2022)Lu, Bartolo, Moore, Riedel, and
  Stenetorp}]{lu2021fantastically}
Yao Lu, Max Bartolo, Alastair Moore, Sebastian Riedel, and Pontus Stenetorp.
  2022.
\newblock \href {https://doi.org/10.18653/v1/2022.acl-long.556} {Fantastically
  ordered prompts and where to find them: Overcoming few-shot prompt order
  sensitivity}.
\newblock In \emph{Proceedings of the 60th Annual Meeting of the Association
  for Computational Linguistics (Volume 1: Long Papers)}, pages 8086--8098,
  Dublin, Ireland. Association for Computational Linguistics.

\bibitem[{McCoy et~al.(2019)McCoy, Pavlick, and Linzen}]{mccoy2019right}
Tom McCoy, Ellie Pavlick, and Tal Linzen. 2019.
\newblock \href {https://doi.org/10.18653/v1/P19-1334} {Right for the wrong
  reasons: Diagnosing syntactic heuristics in natural language inference}.
\newblock In \emph{Proceedings of the 57th Annual Meeting of the Association
  for Computational Linguistics}, pages 3428--3448, Florence, Italy.
  Association for Computational Linguistics.

\bibitem[{Miller et~al.(2020)Miller, Krauth, Recht, and
  Schmidt}]{miller2020effect}
John Miller, Karl Krauth, Benjamin Recht, and Ludwig Schmidt. 2020.
\newblock \href {http://proceedings.mlr.press/v119/miller20a.html} {The effect
  of natural distribution shift on question answering models}.
\newblock In \emph{Proceedings of the 37th International Conference on Machine
  Learning, {ICML} 2020, 13-18 July 2020, Virtual Event}, volume 119 of
  \emph{Proceedings of Machine Learning Research}, pages 6905--6916. {PMLR}.

\bibitem[{Min et~al.(2022)Min, Lyu, Holtzman, Artetxe, Lewis, Hajishirzi, and
  Zettlemoyer}]{min2022rethinking}
Sewon Min, Xinxi Lyu, Ari Holtzman, Mikel Artetxe, Mike Lewis, Hannaneh
  Hajishirzi, and Luke Zettlemoyer. 2022.
\newblock \href {https://aclanthology.org/2022.emnlp-main.759} {Rethinking the
  role of demonstrations: What makes in-context learning work?}
\newblock In \emph{Proceedings of the 2022 Conference on Empirical Methods in
  Natural Language Processing}, pages 11048--11064, Abu Dhabi, United Arab
  Emirates. Association for Computational Linguistics.

\bibitem[{Mishra et~al.(2022)Mishra, Khashabi, Baral, Choi, and
  Hajishirzi}]{mishra2021reframing}
Swaroop Mishra, Daniel Khashabi, Chitta Baral, Yejin Choi, and Hannaneh
  Hajishirzi. 2022.
\newblock \href {https://doi.org/10.18653/v1/2022.findings-acl.50} {Reframing
  instructional prompts to {GPT}k{'}s language}.
\newblock In \emph{Findings of the Association for Computational Linguistics:
  ACL 2022}, pages 589--612, Dublin, Ireland. Association for Computational
  Linguistics.

\bibitem[{Mosbach et~al.(2023)Mosbach, Pimentel, Ravfogel, Klakow, and
  Elazar}]{mosbach2023few}
Marius Mosbach, Tiago Pimentel, Shauli Ravfogel, Dietrich Klakow, and Yanai
  Elazar. 2023.
\newblock \href {https://arxiv.org/abs/2305.16938} {Few-shot fine-tuning vs.
  in-context learning: A fair comparison and evaluation}.
\newblock \emph{ArXiv preprint}, abs/2305.16938.

\bibitem[{Nie et~al.(2020)Nie, Williams, Dinan, Bansal, Weston, and
  Kiela}]{nie2019adversarial}
Yixin Nie, Adina Williams, Emily Dinan, Mohit Bansal, Jason Weston, and Douwe
  Kiela. 2020.
\newblock \href {https://doi.org/10.18653/v1/2020.acl-main.441} {Adversarial
  {NLI}: A new benchmark for natural language understanding}.
\newblock In \emph{Proceedings of the 58th Annual Meeting of the Association
  for Computational Linguistics}, pages 4885--4901, Online. Association for
  Computational Linguistics.

\bibitem[{Niven and Kao(2019)}]{niven2019probing}
Timothy Niven and Hung-Yu Kao. 2019.
\newblock \href {https://doi.org/10.18653/v1/P19-1459} {Probing neural network
  comprehension of natural language arguments}.
\newblock In \emph{Proceedings of the 57th Annual Meeting of the Association
  for Computational Linguistics}, pages 4658--4664, Florence, Italy.
  Association for Computational Linguistics.

\bibitem[{Ohmer et~al.(2023)Ohmer, Bruni, and Hupkes}]{ohmer2023evaluating}
Xenia Ohmer, Elia Bruni, and Dieuwke Hupkes. 2023.
\newblock \href {https://arxiv.org/abs/2305.11662} {Evaluating task
  understanding through multilingual consistency: A chatgpt case study}.
\newblock \emph{ArXiv preprint}, abs/2305.11662.

\bibitem[{Ouyang et~al.(2022)Ouyang, Wu, Jiang, Almeida, Wainwright, Mishkin,
  Zhang, Agarwal, Slama, Ray et~al.}]{ouyang2022training}
Long Ouyang, Jeffrey Wu, Xu~Jiang, Diogo Almeida, Carroll Wainwright, Pamela
  Mishkin, Chong Zhang, Sandhini Agarwal, Katarina Slama, Alex Ray, et~al.
  2022.
\newblock Training language models to follow instructions with human feedback.
\newblock \emph{Advances in Neural Information Processing Systems},
  35:27730--27744.

\bibitem[{Poliak et~al.(2018)Poliak, Naradowsky, Haldar, Rudinger, and
  Van~Durme}]{poliak2018hypothesis}
Adam Poliak, Jason Naradowsky, Aparajita Haldar, Rachel Rudinger, and Benjamin
  Van~Durme. 2018.
\newblock \href {https://doi.org/10.18653/v1/S18-2023} {Hypothesis only
  baselines in natural language inference}.
\newblock In \emph{Proceedings of the Seventh Joint Conference on Lexical and
  Computational Semantics}, pages 180--191, New Orleans, Louisiana. Association
  for Computational Linguistics.

\bibitem[{Rajpurkar et~al.(2016)Rajpurkar, Zhang, Lopyrev, and
  Liang}]{rajpurkar2016squad}
Pranav Rajpurkar, Jian Zhang, Konstantin Lopyrev, and Percy Liang. 2016.
\newblock \href {https://doi.org/10.18653/v1/D16-1264} {{SQ}u{AD}: 100,000+
  questions for machine comprehension of text}.
\newblock In \emph{Proceedings of the 2016 Conference on Empirical Methods in
  Natural Language Processing}, pages 2383--2392, Austin, Texas. Association
  for Computational Linguistics.

\bibitem[{Sch{\"{o}}lkopf et~al.(2012)Sch{\"{o}}lkopf, Janzing, Peters,
  Sgouritsa, Zhang, and Mooij}]{scholkopf2012causal}
Bernhard Sch{\"{o}}lkopf, Dominik Janzing, Jonas Peters, Eleni Sgouritsa, Kun
  Zhang, and Joris~M. Mooij. 2012.
\newblock \href {http://icml.cc/2012/papers/625.pdf} {On causal and anticausal
  learning}.
\newblock In \emph{Proceedings of the 29th International Conference on Machine
  Learning, {ICML} 2012, Edinburgh, Scotland, UK, June 26 - July 1, 2012}.
  icml.cc / Omnipress.

\bibitem[{Si et~al.(2023)Si, Gan, Yang, Wang, Wang, Boyd-Graber, and
  Wang}]{si2023prompting}
Chenglei Si, Zhe Gan, Zhengyuan Yang, Shuohang Wang, Jianfeng Wang, Jordan
  Boyd-Graber, and Lijuan Wang. 2023.
\newblock \href {http://arxiv.org/abs/2210.09150} {Prompting gpt-3 to be
  reliable}.

\bibitem[{Taori et~al.(2023)Taori, Gulrajani, Zhang, Dubois, Li, Guestrin,
  Liang, and Hashimoto}]{alpaca}
Rohan Taori, Ishaan Gulrajani, Tianyi Zhang, Yann Dubois, Xuechen Li, Carlos
  Guestrin, Percy Liang, and Tatsunori~B. Hashimoto. 2023.
\newblock Stanford alpaca: An instruction-following llama model.
\newblock \url{https://github.com/tatsu-lab/stanford_alpaca}.

\bibitem[{Touvron et~al.(2023)Touvron, Lavril, Izacard, Martinet, Lachaux,
  Lacroix, Rozi{\`e}re, Goyal, Hambro, Azhar et~al.}]{touvron2023llama}
Hugo Touvron, Thibaut Lavril, Gautier Izacard, Xavier Martinet, Marie-Anne
  Lachaux, Timoth{\'e}e Lacroix, Baptiste Rozi{\`e}re, Naman Goyal, Eric
  Hambro, Faisal Azhar, et~al. 2023.
\newblock \href {https://arxiv.org/abs/2302.13971} {Llama: Open and efficient
  foundation language models}.
\newblock \emph{ArXiv preprint}, abs/2302.13971.

\bibitem[{Wang et~al.(2022)Wang, Kordi, Mishra, Liu, Smith, Khashabi, and
  Hajishirzi}]{wang2022self}
Yizhong Wang, Yeganeh Kordi, Swaroop Mishra, Alisa Liu, Noah~A Smith, Daniel
  Khashabi, and Hannaneh Hajishirzi. 2022.
\newblock \href {https://arxiv.org/abs/2212.10560} {Self-instruct: Aligning
  language model with self generated instructions}.
\newblock \emph{ArXiv preprint}, abs/2212.10560.

\bibitem[{Wang et~al.(2017)Wang, Hamza, and Florian}]{wang2017bilateral}
Zhiguo Wang, Wael Hamza, and Radu Florian. 2017.
\newblock \href {https://doi.org/10.24963/ijcai.2017/579} {Bilateral
  multi-perspective matching for natural language sentences}.
\newblock In \emph{Proceedings of the Twenty-Sixth International Joint
  Conference on Artificial Intelligence, {IJCAI} 2017, Melbourne, Australia,
  August 19-25, 2017}, pages 4144--4150. ijcai.org.

\bibitem[{Wei et~al.(2022)Wei, Bosma, Zhao, Guu, Yu, Lester, Du, Dai, and
  Le}]{wei2021finetuned}
Jason Wei, Maarten Bosma, Vincent~Y. Zhao, Kelvin Guu, Adams~Wei Yu, Brian
  Lester, Nan Du, Andrew~M. Dai, and Quoc~V. Le. 2022.
\newblock \href {https://openreview.net/forum?id=gEZrGCozdqR} {Finetuned
  language models are zero-shot learners}.
\newblock In \emph{The Tenth International Conference on Learning
  Representations, {ICLR} 2022, Virtual Event, April 25-29, 2022}.
  OpenReview.net.

\bibitem[{Wei et~al.(2023)Wei, Wei, Tay, Tran, Webson, Lu, Chen, Liu, Huang,
  Zhou et~al.}]{wei2023larger}
Jerry Wei, Jason Wei, Yi~Tay, Dustin Tran, Albert Webson, Yifeng Lu, Xinyun
  Chen, Hanxiao Liu, Da~Huang, Denny Zhou, et~al. 2023.
\newblock \href {https://arxiv.org/abs/2303.03846} {Larger language models do
  in-context learning differently}.
\newblock \emph{ArXiv preprint}, abs/2303.03846.

\bibitem[{Williams et~al.(2018)Williams, Nangia, and
  Bowman}]{williams2017broad}
Adina Williams, Nikita Nangia, and Samuel Bowman. 2018.
\newblock \href {https://doi.org/10.18653/v1/N18-1101} {A broad-coverage
  challenge corpus for sentence understanding through inference}.
\newblock In \emph{Proceedings of the 2018 Conference of the North {A}merican
  Chapter of the Association for Computational Linguistics: Human Language
  Technologies, Volume 1 (Long Papers)}, pages 1112--1122, New Orleans,
  Louisiana. Association for Computational Linguistics.

\bibitem[{Wolf et~al.(2019)Wolf, Debut, Sanh, Chaumond, Delangue, Moi, Cistac,
  Rault, Louf, Funtowicz et~al.}]{wolf2019huggingface}
Thomas Wolf, Lysandre Debut, Victor Sanh, Julien Chaumond, Clement Delangue,
  Anthony Moi, Pierric Cistac, Tim Rault, R{\'e}mi Louf, Morgan Funtowicz,
  et~al. 2019.
\newblock \href {https://arxiv.org/abs/1910.03771} {Huggingface's transformers:
  State-of-the-art natural language processing}.
\newblock \emph{ArXiv preprint}, abs/1910.03771.

\bibitem[{Yoo et~al.(2022)Yoo, Kim, Kim, Cho, Jo, Lee, Lee, and
  Kim}]{kim2022ground}
Kang~Min Yoo, Junyeob Kim, Hyuhng~Joon Kim, Hyunsoo Cho, Hwiyeol Jo, Sang-Woo
  Lee, Sang-goo Lee, and Taeuk Kim. 2022.
\newblock \href {https://aclanthology.org/2022.emnlp-main.155} {Ground-truth
  labels matter: A deeper look into input-label demonstrations}.
\newblock In \emph{Proceedings of the 2022 Conference on Empirical Methods in
  Natural Language Processing}, pages 2422--2437, Abu Dhabi, United Arab
  Emirates. Association for Computational Linguistics.

\bibitem[{Zhang et~al.(2019)Zhang, Baldridge, and He}]{zhang2019paws}
Yuan Zhang, Jason Baldridge, and Luheng He. 2019.
\newblock \href {https://doi.org/10.18653/v1/N19-1131} {{PAWS}: Paraphrase
  adversaries from word scrambling}.
\newblock In \emph{Proceedings of the 2019 Conference of the North {A}merican
  Chapter of the Association for Computational Linguistics: Human Language
  Technologies, Volume 1 (Long and Short Papers)}, pages 1298--1308,
  Minneapolis, Minnesota. Association for Computational Linguistics.

\bibitem[{Zhao et~al.(2021)Zhao, Wallace, Feng, Klein, and
  Singh}]{zhao2021calibrate}
Zihao Zhao, Eric Wallace, Shi Feng, Dan Klein, and Sameer Singh. 2021.
\newblock \href {http://proceedings.mlr.press/v139/zhao21c.html} {Calibrate
  before use: Improving few-shot performance of language models}.
\newblock In \emph{Proceedings of the 38th International Conference on Machine
  Learning, {ICML} 2021, 18-24 July 2021, Virtual Event}, volume 139 of
  \emph{Proceedings of Machine Learning Research}, pages 12697--12706. {PMLR}.

\bibitem[{Zhong et~al.(2021)Zhong, Lee, Zhang, and Klein}]{zhong2021adapting}
Ruiqi Zhong, Kristy Lee, Zheng Zhang, and Dan Klein. 2021.
\newblock \href {https://doi.org/10.18653/v1/2021.findings-emnlp.244} {Adapting
  language models for zero-shot learning by meta-tuning on dataset and prompt
  collections}.
\newblock In \emph{Findings of the Association for Computational Linguistics:
  EMNLP 2021}, pages 2856--2878, Punta Cana, Dominican Republic. Association
  for Computational Linguistics.

\bibitem[{Zhou et~al.(2023)Zhou, Liu, Xu, Iyer, Sun, Mao, Ma, Efrat, Yu, Yu,
  Zhang, Ghosh, Lewis, Zettlemoyer, and Levy}]{zhou2023lima}
Chunting Zhou, Pengfei Liu, Puxin Xu, Srini Iyer, Jiao Sun, Yuning Mao, Xuezhe
  Ma, Avia Efrat, Ping Yu, Lili Yu, Susan Zhang, Gargi Ghosh, Mike Lewis, Luke
  Zettlemoyer, and Omer Levy. 2023.
\newblock \href {http://arxiv.org/abs/2305.11206} {Lima: Less is more for
  alignment}.

\end{thebibliography}
\bibliographystyle{acl_natbib}

\onecolumn
\appendix
\section{Experiment 1: List of TT models}
\label{app:used_models}
We compare the sensitivity to spurious correlations of ICL models with TT models. The following table contains all TT models we used during these experiments, providing the respective handle for the huggingface hub or indicating with `own' that we fine-tuned the respective model ourselves. \\

\begin{tabular}{l|l|c|c|c}
\multicolumn{2}{c}{}&\multicolumn{2}{c}{Models}&\\
\cline{3-4}
\multicolumn{2}{c|}{}&RoBERTa\textsubscript{BASE}&RoBERTa\textsubscript{LARGE}\\ 
\cline{2-4}
& MNLI & textattack/roberta-base-MNLI & roberta-large-mnli \\ 
\cline{2-4}
Base datasets & SQuAD & deepset/roberta-base-squad2 & deepset/roberta-large-squad2 \\ 
\cline{2-4}
& QQP & own & own \\ 
\cline{2-4}
\multicolumn{2}{c}{}&\multicolumn{2}{c}{}&\\
\cline{2-4}
& HANS & own & own \\ 
\cline{2-4}
& ANLI  & own & own \\ 
\cline{2-4}
\multirow{2}{*}{Adv. datasets} & PAWS & own & own \\ 
\cline{2-4}
& SQuAD adversarial & own & own \\ 
\cline{2-4}
& adversarial QA & own & own \\ 
\cline{2-4}
& SQuAD shifts & own & own \\ 
\cline{2-4}
\end{tabular}
\vspace{0.5cm}

\section{Experiment 1: Finetuning details of own models}
\label{app:ft_details}

We finetuned all RoBERTa models using the same set of hyperparameters, based on the literature and experience.
\paragraph{Hyperparameters}
We train using the ADAM Optimizer with $\gamma$ = 1e-05, inverse square root decay and $\beta_{1/2}$ = (0.9, 0.999), no weight decay, 250 warmup steps and a batch size of 8.
We stop training if the model does not show improvement on the validation set for 1 epoch of training.

\paragraph{Data} 
For adversarially tuned models, we mixed the training set of the base data with 70\% of the adversarial data (30\% retained for evaluation). 
We ensured a mixing ratio of 20\%/80\% adversarial/base data.

\section{Experiment 1: Datasets details}
\label{app:dataset_details}

We here provide additional information about the datasets we use in Experiment 1:

\subsection{Base datasets}
\label{app:dataset_details:base}

\begin{description}[style=nextline]
    \item[MNLI (Multi Natural Language inference; \citealt{williams2017broad})] 
    A large-scale natural language inference dataset. It contains sentence pairs annotated with three categories: entailment, contradiction, and neutral. The dataset is sourced from a variety of genres, like fiction, government documents, and telephone conversations, thus encouraging models to learn domain-agnostic representations.

    \item[QQP (Quora Question Pairs; \citealt{wang2017bilateral})]
    A collection of question pairs from the Quora platform, labelled as either duplicates or non-duplicates. The aim is to identify semantically equivalent questions, addressing challenges such as paraphrasing and varying levels of detail.

    \item[SQuAD (Stanford Question Answering Dataset; \citealt{rajpurkar2016squad})]
    A reading comprehension dataset consisting of questions about passages from Wikipedia. The questions are human-annotated, and the answer to each question is a segment (or span) of the passage. The goal of models is to identify and extract the correct span from the passage that answers the question.
\end{description}

\subsection{Adversarial datasets}
\label{app:dataset_details:adversarial}

\begin{description}[style=nextline]
    \item[HANS (Heuristic Analysis for NLI Systems; \citealt{mccoy2019right})] 
    Constructed to evaluate models on non-entailment cases that appear entailed due to spurious biases. Built upon common NLI datasets like SNLI and MultiNLI, it dissects three heuristic strategies that a model might utilise: lexical overlap, subsequence, and syntactic structure.

    \item[ANLI (Adversarial Natural Language Inference; \citealt{nie2019adversarial})]
     Generated by first training models on existing datasets (e.g., SNLI and MultiNLI) and then having human annotators produce examples that the models predict incorrectly. Generation of additional examples was done in multiple rounds with respectively improved models, accordingly each round increases the adversarial difficulty.

    \item[PAWS (Paraphrase Adversaries from Word Scrambling; \citealt{zhang2019paws})]
    Comprises sentence pairs with high lexical overlap but differing semantics, challenging models that heavily weigh word overlap. An adversarial expansion to datasets like the Quora Question Pairs dataset (QQP).

    \item[SQuAD Adversarial \citep{jia2017adversarial}]
    A derivative of the Stanford Question Answering Dataset (SQuAD) where adversarial sentences are introduced into the context paragraphs, aiming to mislead models into selecting incorrect answers while the correct answers remain unchanged.

    \item[Adversarial QA \citep{bartolo2020beat}]
    A reading comprehension dataset, where each question is tied to a Wikipedia passage. Distinctively, answer annotations are freeform human responses rather than extracts from the passage, testing the extractive capability boundaries of SQuAD-inspired models.

    \item[SQuAD Shifts \citep{miller2020effect}]
    Formed by perturbing the original SQuAD distribution in terms of linguistic and stylistic attributes. This dataset gauges model robustness against unseen data distributions, such as domain shifts or synthetic noise.

\end{description}

\section{Experiment 1: Impact of spurious correlations in ICL}
\label{app:ANOVA_details}

We conducted an additional analysis of the results in  Section~\ref{subsec:exp1_results}. 
The goal of this additional analysis is to understand the impact of the type of adaptation data (adversarial vs. base) on the prediction outcomes in comparison with \textit{other} factors that we varied in our experiments (such as the type of \factor{instruction template}, whether the model was \factor{instruction tuned} or the \factor{size} of the model).
\factor{Type data} is a binary factor indicating whether the model was adapted on base or adversarial data; \factor{Size} is a quarternary factor indicating model size; \factor{Type instructions} is a binary factor indicating the type of template that was used; \factor{Instruction tuned} is a binary factor indicating whether the tested model was instruction tuned or not.

Table~\ref{tab:anova_details} shows the summary statistics of an ANOVA that we apply to these factors and their impact on the model accuracy.
We can see from Table~\ref{tab:anova_details} that adaptation data is the only factor that does not significantly impact prediction outcomes.

\begin{table}[h]
\centering

\begin{tabular}{llllll}
\toprule
 & \textbf{df} & \textbf{sum\_sq} & \textbf{mean\_sq} & \textbf{F} & \textbf{PR(>F)}  \\
\midrule
Type data & 1.0 & 8.67 & 8.67 & 0.12 & 0.72 \\
Size & 3.0 & 6626.73 & 2208.91 & 31.26 & 5.71e-18 \\
Type instruction & 1.0 & 95.32 & 95.32 & 1.34 & 0.024 \\
Instruction tuned & 1.0 & 900.55 & 900.55 & 12.74 & 4.05e-04 \\
Residual & 357.0 & 25220.11 & 70.64 & NaN & NaN \\
\bottomrule
\end{tabular}
\caption{Results of ANOVA}
\label{tab:anova_details}
\end{table}

\section{Experiment 1 \& 2:Prompt template examples}
\label{app:template_examples}

\subsection{FLAN instructions}
\label{subapp:flan}
Input: \\
\fbox{\parbox{\linewidth}{
Does the Hypothesis in the input entail (True) or contradict (False) the Premise or is it independent (Neither)?\\
Premise: Kirklees Stadium (known as the John Smith's Stadium due to sponsorship), is a multi-use sports stadium in Huddersfield in West Yorkshire, England. Since 1994, it has been the home ground of football club Huddersfield Town and rugby league side Huddersfield Giants, both of whom moved from Leeds Road.\\
Hypothesis: Kirklees Stadium is in Scotland. \\

OPTIONS: \\
- True  \\
- Neither \\
- False \\

ANSWER: False. \\

\textbf{[...]} \\

Does the Hypothesis in the input entail (True) or contradict (False) the Premise or is it independent (Neither)?\\
Premise: Jonathan Smith (born January 17, 1971), better known by his stage name Lil Jon, is an American rapper, record producer, and DJ. He was the frontman of the group Lil Jon \& The East Side Boyz, which he formed in 1997, and they released several albums until 2004.\\
Hypothesis: Jonathan Smith spent much of his time in China. \\

OPTIONS: \\
- True  \\
- Neither \\
- False \\

ANSWER: 
}}

\vspace{10pt}
\noindent Target: \\
\fbox{\parbox{\linewidth}{Neither.
}}

\subsection{P3 details}
\label{subapp:p3}

In the following, we provide more details on the instruction templates \citep{bach2022promptsource}, as used in Experiments II.

\subsubsection{P3 details -- Names}
\label{subsubapp:p3_names}
Names of all available P3-instructions, following the ordering of Figure~\ref{fig:diversity_predictions}:
\begin{multicols}{3}
\begin{enumerate}
\small
\setlength\itemsep{0.1em}
\item `MNLI Crowdsource'
\item `Guaranteed Possible Impossible'
\item `Always Sometimes Never'
\item `Consider Always Sometimes Never'
\item `Does This Imply'
\item `Guaranteed True'
\item `GPT 3 Style'
\item `Take the Following as Truth'
\item `Must Be True'
\item `Based on the Previous Passage'
\item `Should Assume'
\item `Can We Infer'
\item `Justified in Saying'
\item `Does It Follow That'
\item `Claim True False Inconclusive'
\end{enumerate}
\end{multicols}

\subsubsection{P3 details -- Examples}
\label{subsubapp:p3_examples}
We here show examples of P3 prompt templates as they are used in Experiment 2: The prompt templates wrap the respective ANLI data point and provide natural language instructions about the task to the model. 

\paragraph{High-performing templates}
`Claim true false inconclusive' \\
\fbox{\parbox{\linewidth}{
\textbf{[...]} \\

Jonathan Smith (born January 17, 1971), better known by his stage name Lil Jon, is an American rapper, record producer, and DJ. He was the frontman of the group Lil Jon \& The East Side Boyz, which he formed in 1997, and they released several albums until 2004. Based on that information, is the claim: "Jonathan Smith spent much of his time in China." true, false, or inconclusive? \\

ANSWER: 
}}
\paragraph{High-performing templates} `Does it follow that' \\
\fbox{\parbox{\linewidth}{
\textbf{[...]} \\
Given that Jonathan Smith (born January 17, 1971), better known by his stage name Lil Jon, is an American rapper, record producer, and DJ. He was the frontman of the group Lil Jon \& The East Side Boyz, which he formed in 1997, and they released several albums until 2004. Does it follow that Jonathan Smith spent much of his time in China. Yes, no, or maybe? \\

ANSWER: 
}}
\paragraph{Low-performing templates} `MNLI crowdsource' \\
\fbox{\parbox{\linewidth}{
\textbf{[...]} \\
Jonathan Smith (born January 17, 1971), better known by his stage name Lil Jon, is an American rapper, record producer, and DJ. He was the frontman of the group Lil Jon \& The East Side Boyz, which he formed in 1997, and they released several albums until 2004. Using only the above description and what you know about the world, "Jonathan Smith spent much of his time in China." is definitely correct, incorrect, or inconclusive? 

ANSWER: 
}}

\paragraph{Low-performing templates}
`Guaranteed possible impossible' \\
\fbox{\parbox{\linewidth}{
\textbf{[...]} \\
Assume it is true that Jonathan Smith (born January 17, 1971), better known by his stage name Lil Jon, is an American rapper, record producer, and DJ. He was the frontman of the group Lil Jon \& The East Side Boyz, which he formed in 1997, and they released several albums until 2004. \\

Therefore, "Jonathan Smith spent much of his time in China." is guaranteed, possible, or impossible? \\

ANSWER: 
}}

\section{Experiment 2: Factors details}
\label{app:factors_details}
In the following, we provide a more detailed description of the factors used in Section~\ref{sec:holistic_eval} and also provide our motivation to include these factors.
\subsection{Invariance factors}
\label{app:factor_details:v_condtions}

\paragraph{Size} We consider models of different sizes.
Model size has been shown to be an important moderating factor in probably all previous studies on in-context learning.

\paragraph{Instruction tuning} We have seen previously that instruction tuning improves the consistency of a model across templates (see Section~\ref{subsubsec:probing_templates}). 
We introduce it as a factor to show which other invariance factors it may affect.

\begin{figure}[h!]
  \centering
    \includegraphics[width=0.7\linewidth]{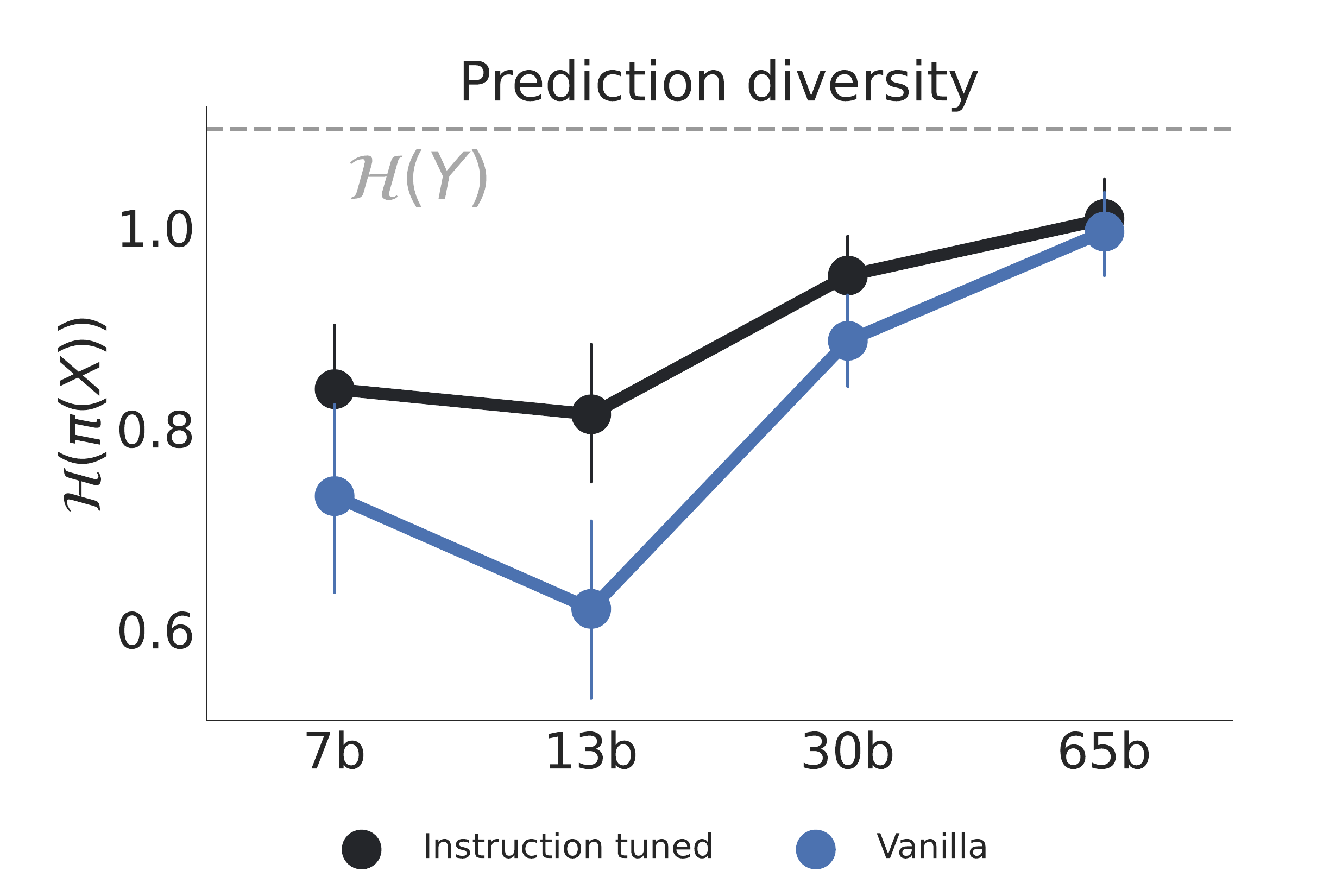}
  \caption{}
  \label{fig:entropy_label_bias}
\end{figure}

\paragraph{Calibration} Previous research has shown how small models are especially biased towards single labels when prompted.
We find similar tendencies for our model:
We exploratively calculate the entropy of a model's predictions across all data points in a dataset. 
This allows us to estimate whether a model is biased toward predicting a single label (low entropy). 
Optimally, a model's prediction should be close to the entropy of the target distribution $\mathcal{H}(Y)$. 
We find that smaller models have a larger bias towards predicting a single label (lower prediction entropy), while larger and IT models get closer to $\mathcal{H}(Y)$ (see Figure~\ref{fig:entropy_label_bias}).

\noindent \citet{zhao2021calibrate} suggests solving this issue by calibrating the model probabilities using `content-free' prompts.
We add the factor of calibration to assess its effects systematically.

\paragraph{n-shots} The number of in-context examples has been shown to interact with other factors
\citep[e.g. according to][\factor{calibration} has a more significant effect for fewer in-context examples]{zhao2021calibrate}. 
We would also expect that \factor{n-shots} interacts with many other in-context factors such as \factor{one label}, in which we show the model just examples with the same label in-context, is modulated by the number of in-context examples.
We introduce `few' ($k$ = 2) and `many' ($k$ = 5) examples as a factor.

\paragraph{Instruction quality} Ultimately, we have seen how some instructions produce consistent and relatively well-performing responses across different models while others do not (see Section~\ref{subsubsec:probing_templates}. 
We add this last factor to see which other types of factors help the in-context learner cope with varying \factor{instruction quality}. 
We chose the two best and two worst-performing templates\footnote{See Appendix~\ref{app:template_examples} for an example of the instructions} from our previous analysis.

\subsection{Invariance factors}
\label{app:factor_details:i_condtions}
The following briefly describes each of the tested $\lambda_{inv}$.

\paragraph{Balanced labels} \citet{zhao2021calibrate} additionally showed how a majority label among the in-context example can influence the distribution of model outputs.
Therefore, we compare contexts with balanced in-context label distribution with randomly sampled labels and an extreme case with only a single in-context label.

\paragraph{Cross-instruction}
\begin{figure*}[h!]
  \centering
    \includegraphics[width=0.7\linewidth]{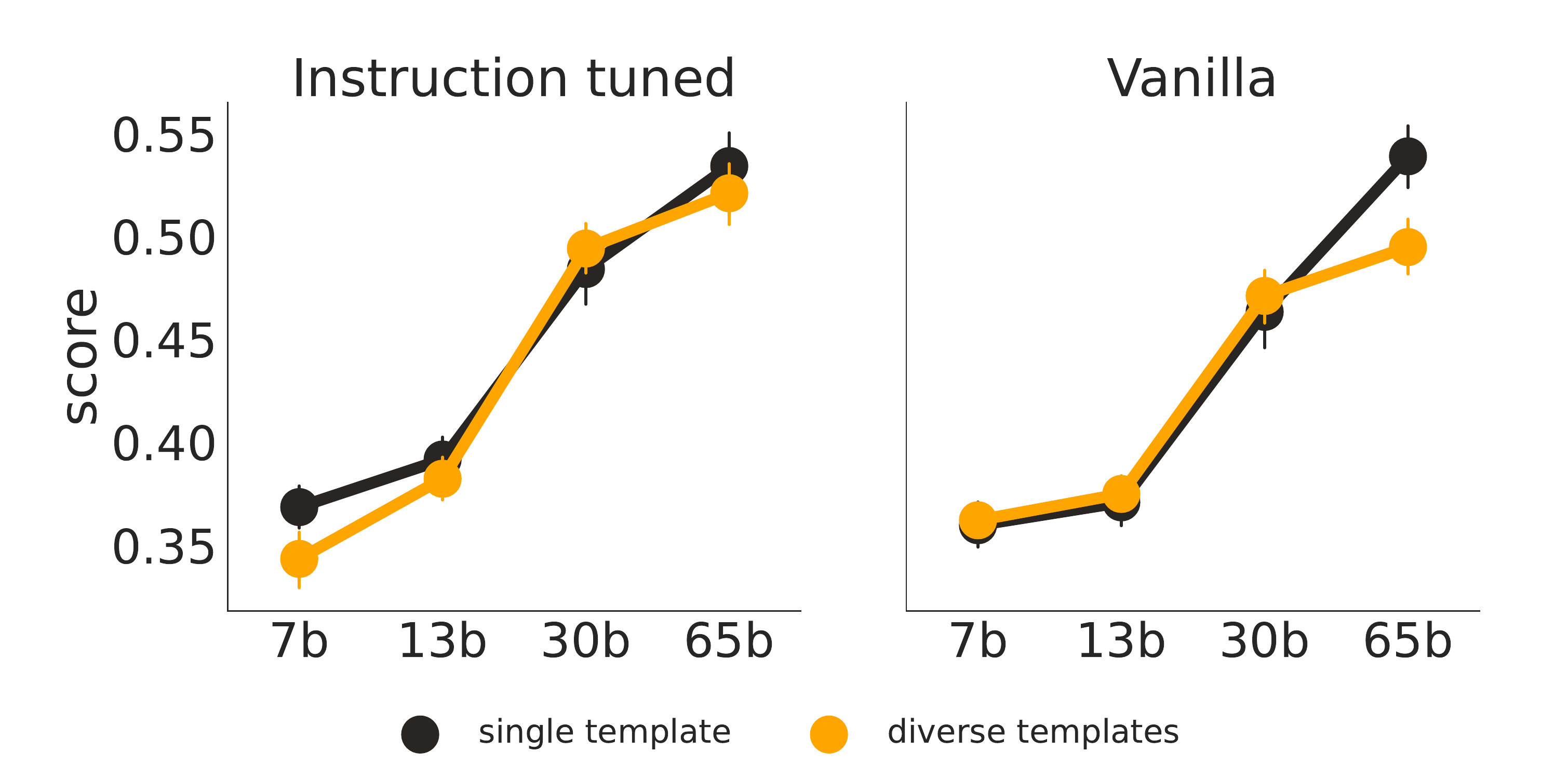}
  \caption{}
  \label{fig:diverse_templates_performance}
\end{figure*}
We include \factor{cross-templates} as a factor to assess model robustness to shifts in label space and surface form of instruction formulation.
Previous research has shown how in-context learners are sensitive to the instructions \citep{mishra2021reframing} as well as the label distribution $\mathcal{C}$ \citep{min2022rethinking}.
The experiments of \citet{min2022rethinking} represent an extreme case in which $\mathcal{C}$ is resampled to be random tokens.
While these edge cases are theoretically attractive, we here change this scenario to a practically common one, where instructions and labels are semantically equivalent but have different surface forms by randomly sampling from the available p3 instructions for the in-context examples. 
We test the impact of in-context instructions in a single setting with results shown in Figure~\ref{fig:diverse_templates_performance}
Surprisingly, almost all models are robust to semantic-invariant changes to instructions of the in-context examples despite changes in the label space and substantial changes in surface form and format across different instructions. 

\paragraph{Cross-task}
\begin{figure*}[h!]
  \centering
    \includegraphics[width=0.7\linewidth]{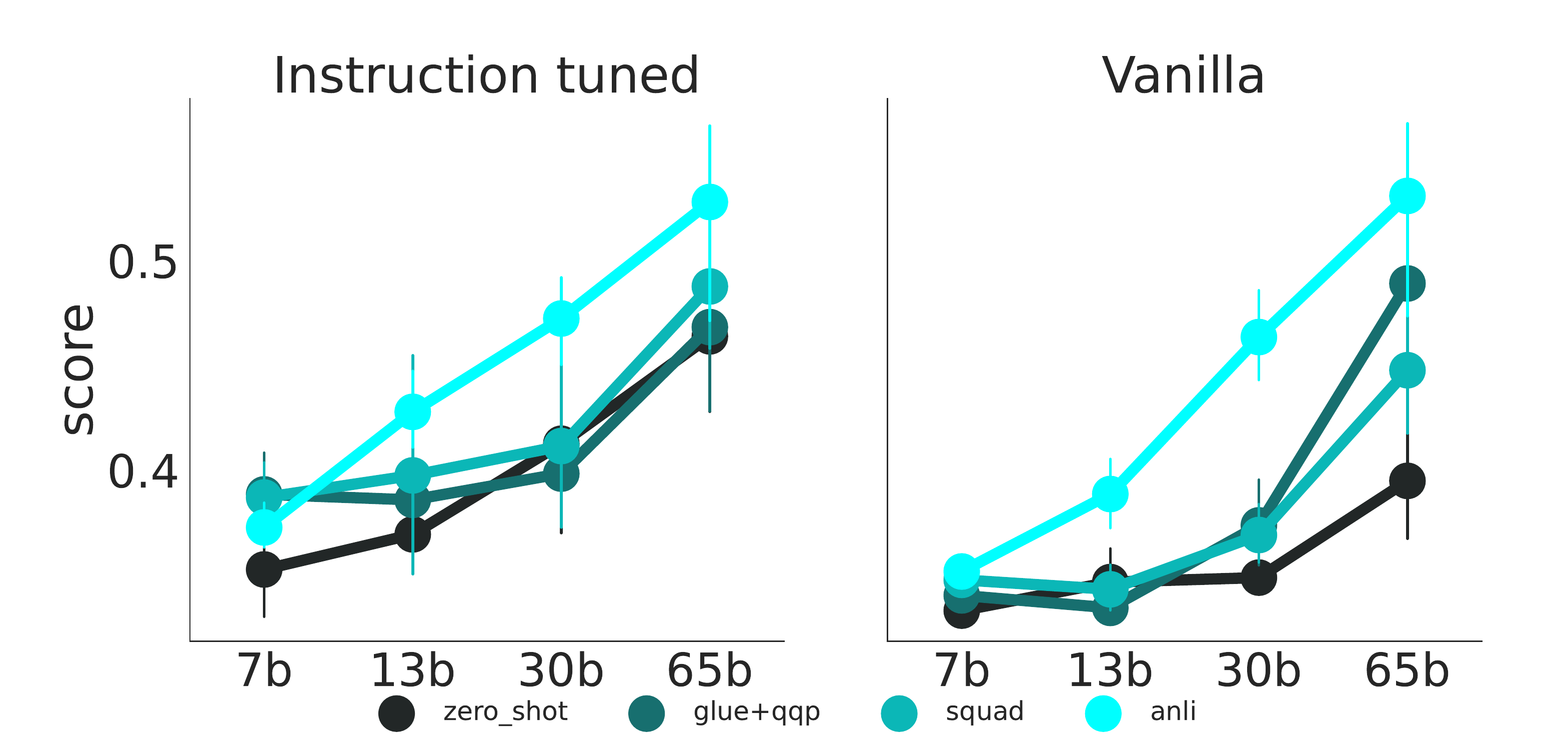}
  \caption{Accuracy scores of all models in all possible setups, with vanilla models on the left and instruction-tuned models on the right.}
  \label{fig:cross_tasks}
\end{figure*}
In \factor{cross-task}, we exchange the task of the in-context examples such that the only consistency between in-context and target examples is the general format ($x$ followed by $y$) and the truthfulness of the $x$ to $y$ mapping.
To see whether conditioning on a fixed label space matters, we add tasks with a discriminative (QQP) and a generative (SQuAD)  objective as different factors.
Compared to a zero-shot baseline, we can see that large models can benefit from conditioning on other tasks (Figure~\ref{fig:cross_tasks}).
For our principal analysis, we only include QQP as an in-context task, as SQuAD is incompatible with many other factors (such as \factor{balanced labels}, \factor{one label} aso...)

\paragraph{Instructions} Besides the quality of the instructions, we are also interested in how consistent model behaviour is across \factor{instructions} that are of similar quality. 
To get an insight into this, we bin the high-quality instructions respectively into a new factor.

\section{Experiment 2: Accuracy distribution}
\label{app:distribution_results}
We here show the distribution of accuracy scores for all setups in experiment 2, separated by model size (hue) and whether the model is instruction tuned or not (i.e. vanilla).

 \begin{figure}[h!]
  \centering
    \includegraphics[width=0.8\linewidth]{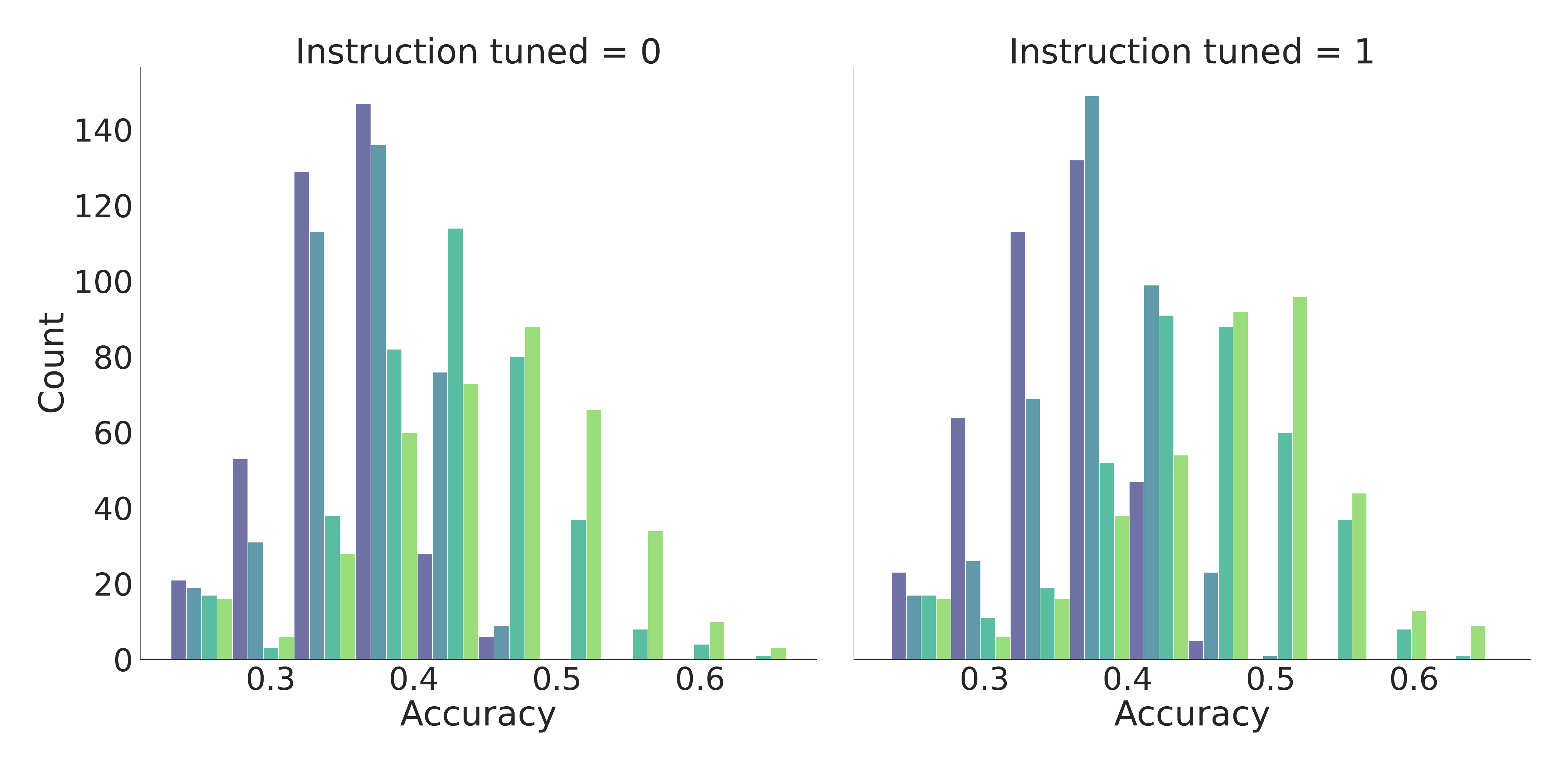}
  \caption{}
  \label{fig:distribution_results}
\end{figure}

\section{Experiment 2: Interactions details}
\label{app:interactions_instructions}

\subsection{ANOVA using \factor{instructions} factor}
We fit an ANOVA using the factor \factor{instructions} instead of \factor{instruction quality}. In that case, we find a similar pattern of interactions, showing that the size of the main effect can not merely explain the number of interactions. 

\begin{figure}[h]
  \begin{subfigure}[h]{0.38\linewidth}
  \centering
    \includegraphics[width=\linewidth]{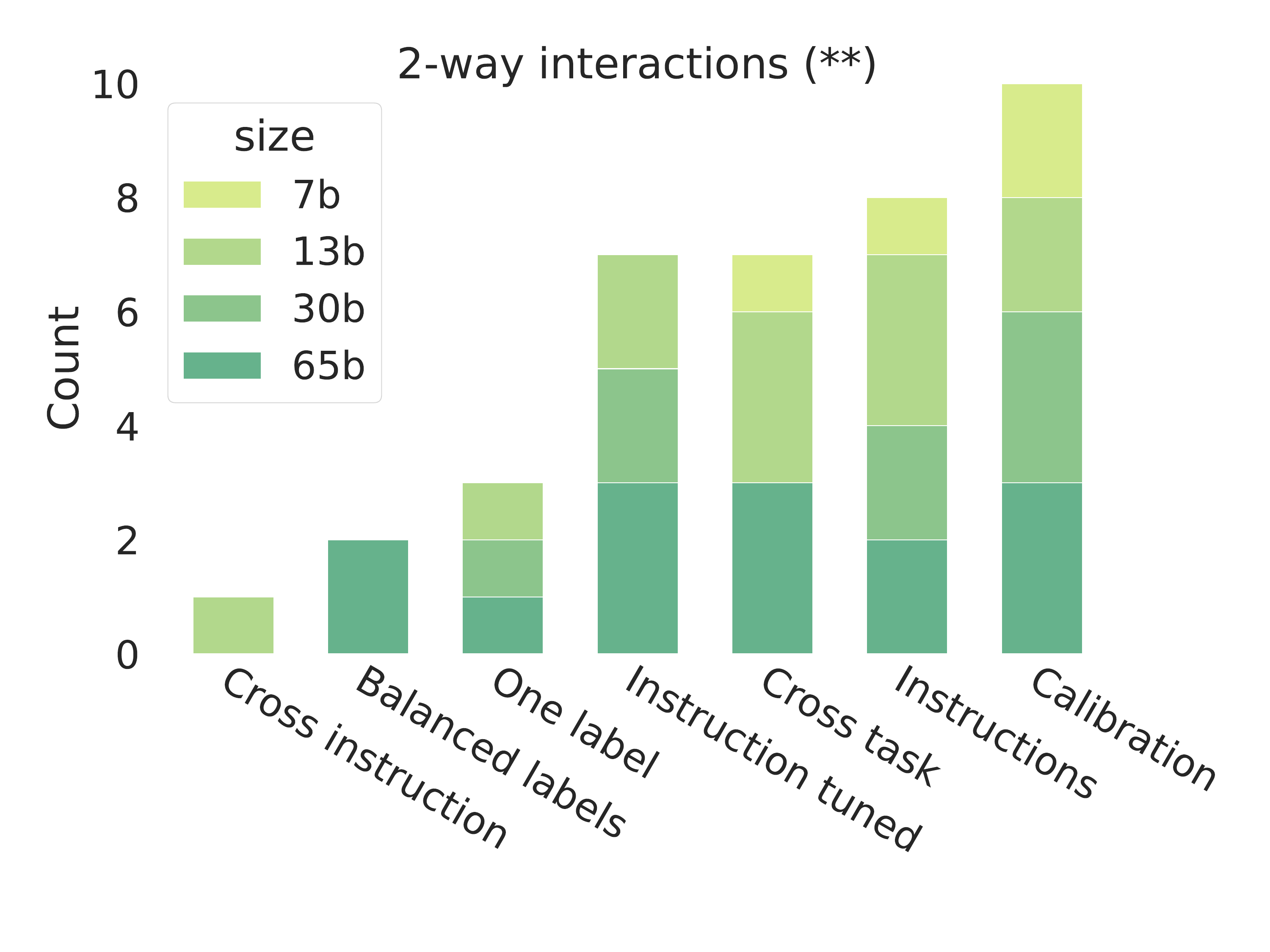}
    \label{subfig:hist_2way_interaction_instructions}
  \end{subfigure}
  \begin{subfigure}[h]{0.38\linewidth}
  \centering
    \includegraphics[width=\linewidth]{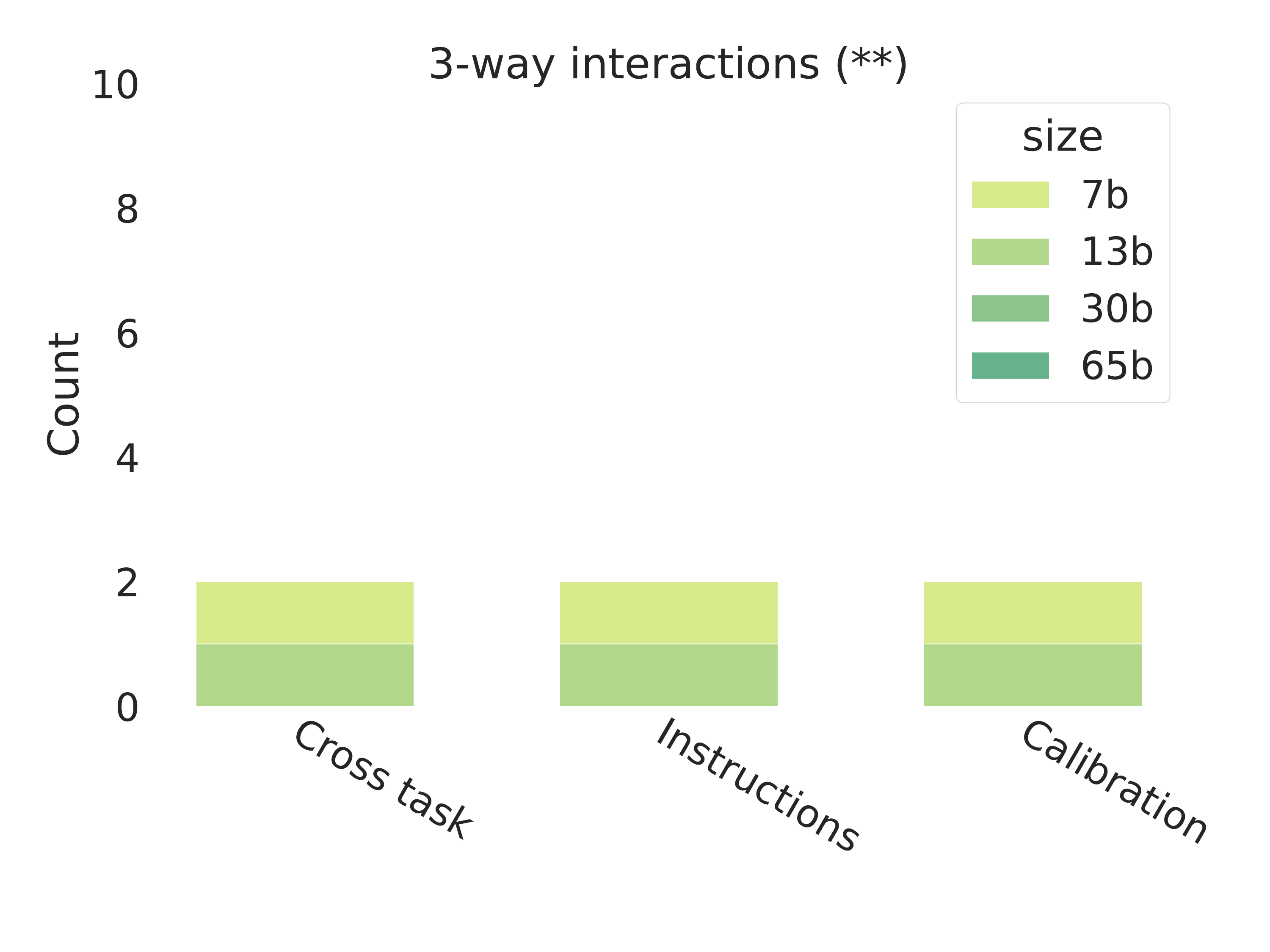}
    \label{subfig:hist_3way_interaction_instructions}
  \end{subfigure}
  \caption{Interactions when excluding \factor{Instruction quality} and keeping \factor{Instructions} instead. We find similar patterns.}
  \label{fig:interactions_instructions}
\end{figure}

\newpage

\subsection{Interaction mappings and effect sizes}
The following shows the exact mapping of the interacting factors as well as the size of the corresponding effect size, measured by $\beta_{\lambda_1\times\lambda_2}$ values from a post hoc regression analysis.

\begin{figure}[h]
  \centering
  \caption{The exact mappings of all 2-way interactions in our experiments.}
    \includegraphics[width=\linewidth]{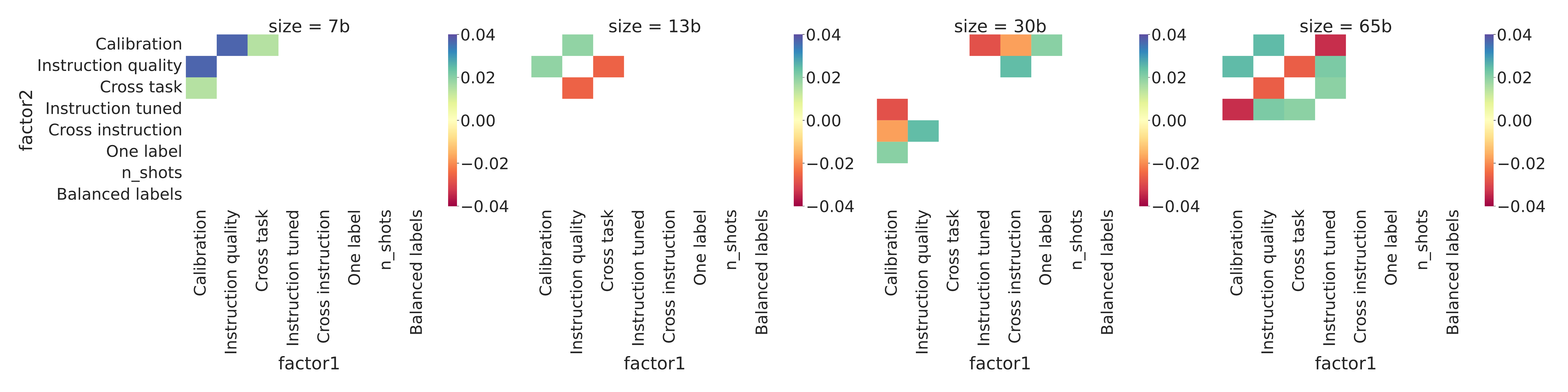}
    \label{fig::heatmap_2way_interaction}
\end{figure}

\begin{table}[htbp]
\centering
\caption{The exact mappings of all 3-way interactions in our experiments.}
\label{tab:instruction_quality}
\begin{tabular}{ccccc}
\hline
\textbf{Model} & \textbf{$\lambda_1$} & \textbf{$\lambda_2$} & \textbf{$\lambda_3$} & \textbf{$\beta_{\lambda_1\times\lambda_2\times\lambda_3}$} \\
\hline
7B & Instruction quality & Calibration & Cross task & 0.037106 \\
13B & Instruction tuned & Calibration & Instruction quality & 0.002102 \\
13B & Instruction quality & Cross task & Calibration & -0.013176 \\
\hline
\end{tabular}
\end{table}

\section{Limitations}
\label{app:limitations}
For the first set of experiments in Section~\ref{sec:spurious_correlations}, the comparison between TT models and ICL is not `fair'. 
Model sizes are not comparable, the amount of adaptation data differs significantly (thousand for task-tuning compared to 5 for ICL) and some of the adversarial datasets were created with some of the TT models `in-the-loop' (e.g. ANLI). 
However, our motivation here is not to be fair, but to show practically relevant effects in either type of task adaptation.
For a fair comparison, see \citet{mosbach2023few}.

For the second set of experiments in Section~\ref{sec:holistic_eval}, we only consider a subset of factors that we deemed the most relevant or interesting. 
Adding more factors would enrich the analysis. 
However, the number of model inferences to compute grows exponentially with the number of considered factors, which sets soft limits for the number of analysed factors.
For potential follow-ups, we suggest a more fine-grained investigation of different instruction designs for the target example, as this potentially yields exciting insights on what exactly leads to the large performance gains and high volatility. 
Our study is coarse in this aspect.

Our analysis would have been more expressive if we chose an `easier' task than the relatively `hard' ANLI dataset to run our evaluation: our smaller models perform relatively poorly across many factors on challenging datasets like ANLI and provide less variance for a meaningful analysis.

\section{Acknowledgements}
\label{app:acknowledgements}

We would like to thank the anonymous reviewers for their time and productive feedback.
Beyond that, we thank the whole COLT group at UPF for their helpful feedback and other support during the project -- especially Emily Cheng and Xixian Liao.
Further, LW thanks the Department of Translation and Language Sciences at the University Pompeu Fabra for funding.

\end{document}